\documentclass[10pt,twocolumn,letterpaper]{article}

\usepackage{iccv}
\usepackage{times}
\usepackage{epsfig}
\usepackage{graphicx}
\usepackage{amsmath}
\usepackage{amssymb}
\usepackage{tabularx}

\usepackage{multirow}

\usepackage[pagebackref=true,breaklinks=true,letterpaper=true,colorlinks,bookmarks=false]{hyperref}

\iccvfinalcopy

\begin{document}

\title{Specifying Object Attributes and Relations in Interactive Scene Generation}

\author{Oron Ashual\\
Tel Aviv University\\
{\tt\small oron.ashual@gmail.com}
\and
Lior Wolf\\
Tel Aviv University and Facebook AI Research\\
{\tt\small wolf@cs.tau.ac.il, wolf@fb.com}
}

\maketitle

\begin{abstract}
We introduce a method for the generation of images from an input scene graph. The method separates between a layout embedding and an appearance embedding. The dual embedding leads to generated images that better match the scene graph, have higher visual quality, and support more complex scene graphs. In addition, the embedding scheme supports multiple and diverse output images per scene graph, which can be further controlled by the user. We demonstrate two modes of per-object control: (i) importing elements from other images, and (ii) navigation in the object space, by selecting an appearance archetype.

Our code is publicly available at \url{https://www.github.com/ashual/scene_generation}.
\end{abstract}

\begin{figure*}
\centering
\includegraphics[width=1.0\linewidth]{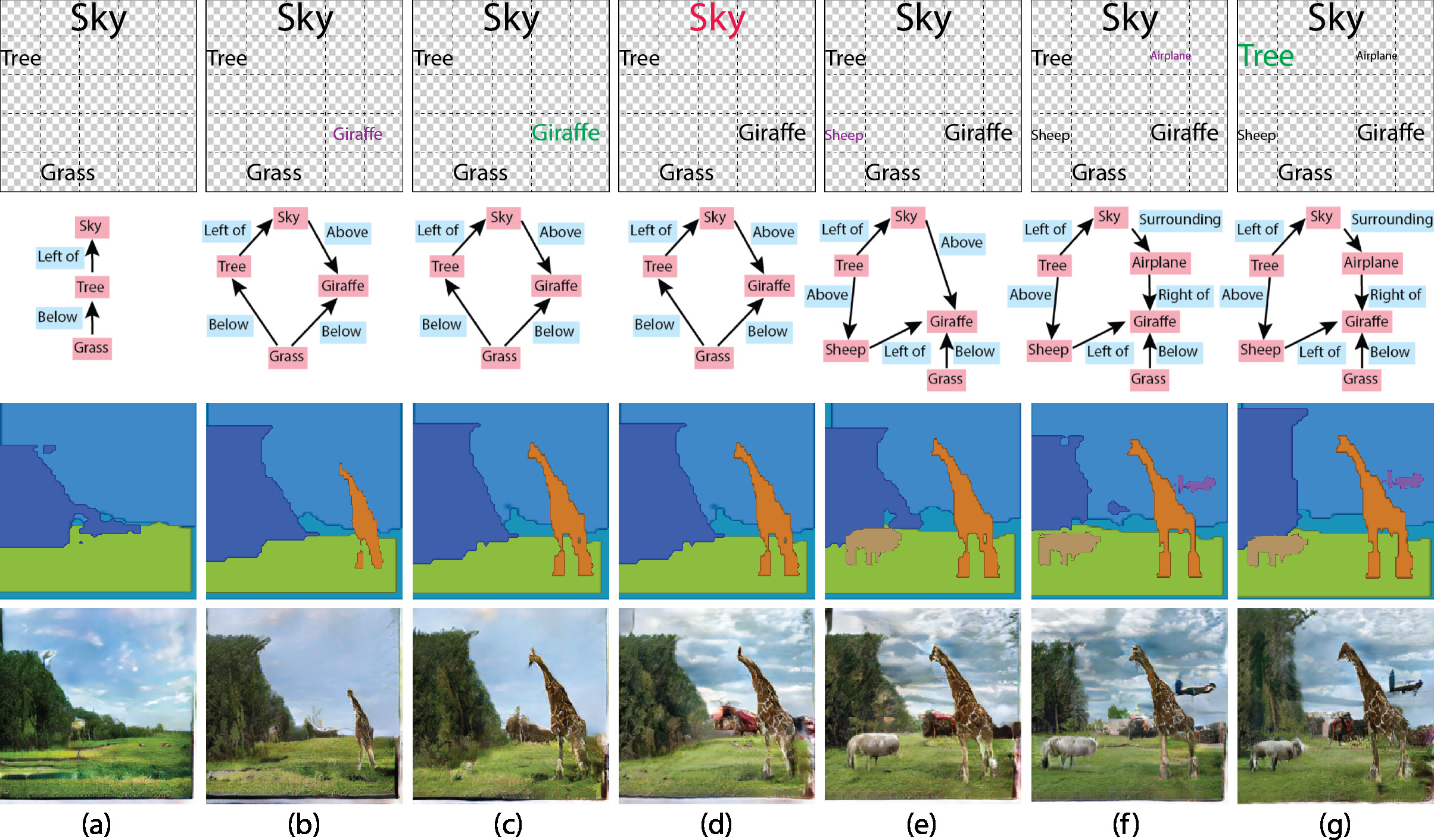}
\caption{An example of the image creation process. (top row) the  schematic illustration panel of the user interface, in which the user arranges the desired objects. (2nd row) the scene graph that is inferred automatically based on this layout. (3rd row) the layout that is created from the scene graph. (bottom row) the generated image. Legend for the GUI colors in the top row: purple -- adding an object, green -- resizing it, red -- replacing its appearance.  (a) A simple layout with a   sky object, a tree and a grass object. All object appearances are initialized to a random archetype appearance. (b) A giraffe is added. (c) The giraffe is enlarged. (d) The appearance of the sky  is changed to a different archetype. (e) A small sheep is added. (f) An airplane is added. (g) The tree is enlarged.}
\label{fig:teaser}
\end{figure*}

\section{Introduction}
David Marr has defined vision as the process of discovering from images what is present in the world, and where it is~\cite{marr}. The combination of {\it what} and {\it where} captures the essence of an image at the semantic level and therefore, also plays a crucial role when defining the desired output of image synthesis tools.

In this work, we employ scene graphs with per-object location and appearance attributes as an accessible and easy-to-manipulate way for users to express their intentions, see Fig.~\ref{fig:teaser}. The {\it what} aspect is captured hierarchically: objects are defined as belonging to a certain class (horse, tree, boat, etc.) and as having certain appearance attributes. These attributes can be (i) selected from a predefined set obtained by clustering previously seen attributes, or (ii) copied from a sample image. The {\it where} aspect, is captured by what is often called a scene graph, i.e., a graph where the scene objects are denoted as nodes, and their relative position, such as ``above'' or ``left of'', are represented as edge types.

Our method employs a dual encoding for each object in the image. The first part encodes the object's placement and captures a relative position and other global image features, as they relate to the specific object. It is generated based on the scene graph, by employing a graph convolution network, followed by the concatenation of a random vector $z$.  The second part encodes the appearance of the object and can be replaced, e.g., by importing it from the same object as it appears in another image, without directly changing the other objects in the image. This copying of objects between images is done in a semantic way, and not at the pixel level.

In the scene graph that we employ, each node is equipped with three types of information: (i) the type of object, encoded as a vector of a fixed dimension, (ii) the location attributes of the objects, which denote the approximate location in the generated image, using a coarse $5\times 5$ grid and its size, discretized to ten values, and (iii) the appearance embedding mentioned above.  The edges denote relations: ``right of'', ``left of'', ``above'', ``below'', ``surrounding'', and ``inside''. The method is implemented within a convenient user interface, which supports a dynamic placement of objects and the creation of a scene graph.
The edge relations are inferred automatically, given the relative position of the objects. This eliminates the need for mostly unnecessary user intervention. Rendering is done in real time, supporting the creation of novel scenes in an interactive way, see Fig.~\ref{fig:teaser}. 

\begin{figure*}[!htb]
  \includegraphics[
  width=1.0\linewidth]{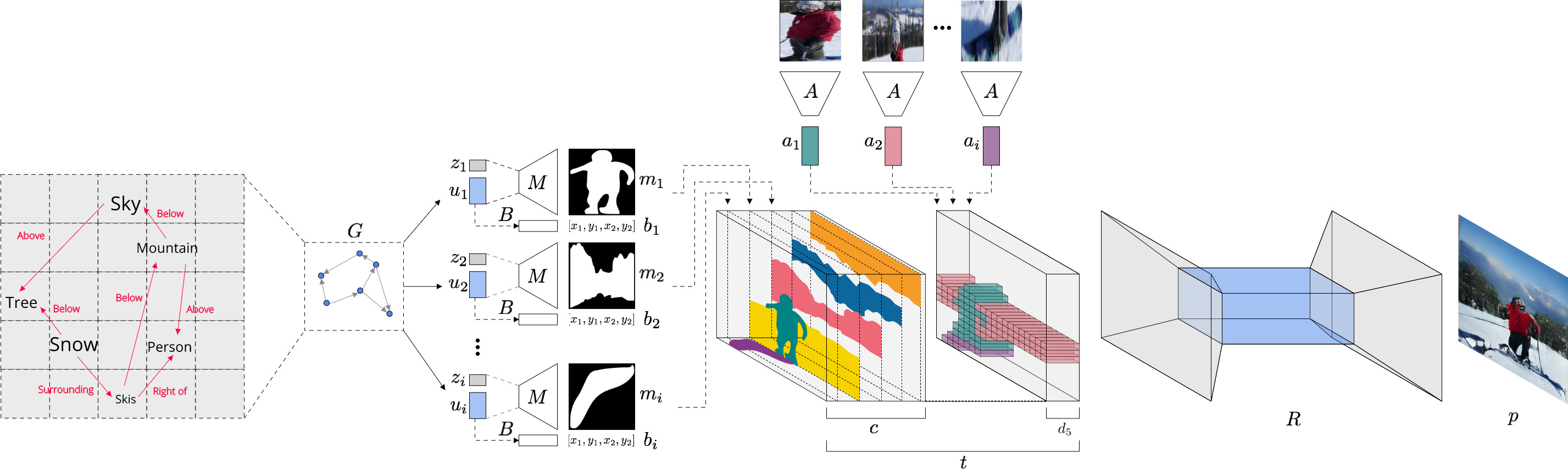}
  \caption{The architecture of our composite network, including the subnetworks $G,M,B,A,R$, and the process of creating the layout tensor $t$. The scene graph is passed to the network $G$ to create the layout embedding $u_i$ of each object. The bounding box $b_i$ is created from this embedding, using network $B$. A random vector $z_i$ is concatenated to $u_i$, and the network $M$ computes the mask $m_i$. The appearance information, as encoded by the network $A$, is then added to create the tensor $t$ with $c + d_5$ channels, $c$ being the number of classes. The autoencoder $R$ generates the final image $p$ from this tensor.}
\label{fig:arch}
\end{figure*}

The neural network that we employ has multiple sub-parts, as can be seen in Fig.~\ref{fig:arch}: (i) A graph convolutional network that converts the input scene graph to a per-object embedding to their location. (ii) A CNN that converts the location embedding of each object to an object's mask. (iii) A parallel network that converts the location embedding to a bounding box location, where the object mask is placed. (iv) An appearance embedding CNN that converts image information into an embedding vector. This process is done off-line and
when creating a new image, the vectors can be imported from other images, or selected from a set of archetypes. (v) A multiplexer that combines the object masks and the appearance embedding information, to create a one multidimensional tensor, where different groups of layers denote different objects. (vi) An encoder-decoder residual network that creates the output image.

Our method is related to the recent work of~\cite{Johnson2018ImageGF}, who create images based on scene graphs. Their method also uses a graph convolutional network to obtain masks, a multiplexer that combines the layout information and a subsequent encoder-decoder architecture for obtaining the final image. There are, however, important differences: (i) by separating the layout embedding from the appearance embedding, we allow for much more control and freedom to the object selection mechanism, (ii) by adding the location attributes as input, we allow for an intuitive and more direct user control, (iii) the architecture we employ enables better quality and higher resolution outputs, (iv) by adding stochasticity before the masks are created, we are able to generate multiple results per scene graph, (v) this effect is amplified by the ability of the users to manipulate the resulting image, by changing the properties of each individual object, (vi) we introduce a mask discriminator, which plays a crucial role in generating plausible masks, (vii) another novel discriminator captures the appearance encoding in a counterfactual way, and (viii) we introduce feature matching based on the discriminator network and (ix) a perceptual loss term to better capture the appearance of an object, even if the pose or shape of that object has changed.

\section{Previous Work}

Image generation techniques based on GANs~\cite{gan} are constantly improving in resolution, visual quality, the diversity of generated images, and the ability to cover the entire visual domain presented during training. In this work, we address conditional image generation, i.e., the creation of images that match a specific input. Earlier work in conditional image generation includes class based image generation~\cite{mirza2014conditional}, which generates an image that matches a given textual description~\cite{RAYLLS16,Zhang2017StackGANTT}. 
In many cases, the conditioning signal is a source image, in which case the problem is often referred to as image translation. Pix2pix~\cite{pix2pix} is a fully supervised method that requires pairs of matching samples from the two domains. The Pix2pixHD architecture that was recently presented by~\cite{wang2018pix2pixHD} is highly influential and many recent video or image mapping works employ elements of it, including our work.

Image generation based on scene graphs was recently presented in~\cite{Johnson2018ImageGF}. A scene graph representation is often used for retrieval based on text~\cite{Johnson2015ImageRU,Quinn2018SemanticIR}, and a few datasets include this information, e.g., COCO-stuff~\cite{Caesar2018COCOStuffTA} and the visual genome~\cite{vg}.
Also related is the synthesis of images from a given input layout of bounding boxes (and not one that is inferred by a network from a scene graph), which was very recently studied by~\cite{Zhao2018ImageGF} for small 64x64 images. In another line of work, images are generated to match input sentences, without constructing the scene graph as an intermediate representation~\cite{RAYLLS16, Zhang2017StackGANTT,semanticLayout}.

Recently, an interactive tool was introduced based on the novel notion of GAN dissection~\cite{bau2018gandissect}. This tool allows for the manipulation of well-localized neurons that control the occurrence of specific objects across the image using a drawing interface. By either adding or reducing the activations of these neurons, objects can be 
added, expanded, reduced or removed.
The manipulation that we offer here is both more semantic (less related to specific locations and more to spatial relations between the objects), and more precise, in the sense that we provide full control over the exact instance of the object and not just of the desired class.

\section{Method}
\label{sec:method}

Each object $i$ in the input scene graph is associated with a single node $n_i=[o_i,l_i]$, where $o_i\in \mathbb{R}^{d1}$ is a learned encoding of the object class and $l_i\in \{0,1\}^{d_2+d_3}$ is a location vector. The object class embedding $o_i$ is one of $c$ possible embedding vectors, $c$ being the number of classes, and $o_i$ is set according to the class of object $i$, denoted $c_i$. The embedding size $d_1$ is set arbitrarily to 128. The first $d_2=25$ bits of $l_i$ denote a coarse image location using a $5\times 5$ grid, and the rest denotes the size of the object, using a scale of $d_3=10$ values. The edge information $e_{ij}\in \mathbb{R}^{d_1}$ exists for a subset of the possible pairs of nodes, and encodes, using a learned embedding, the relations between the nodes. In other words, the values of $e_{ij}$ are taken from a learned dictionary with six possible values, each associated with one type of pairwise relation.

The location of each generated object is given as a pseudo-binary mask $m_i$ (output of a sigmoid) and a bounding box $b_i=[x_1,y_1,x_2,y_2]^\top \in [0,1]^4$, which encodes the coordinates of the bounding box as a ratio of the image dimensions. The mask, but not the bounding box, is also determined by a per-object random vector $z_i \sim  \mathbb{N}(0,1)^{d_4}$ to create a variation in the generated masks, where $d_4=64$ was set arbitrarily, without testing other values.

The method employs multiple ways of embedding input information. The class identity and every inter-object relation, both taking a discrete value, are captured by embedding vectors of dimension $d_1$, which are learned as part of the end-to-end training. The object appearance $a_i\in \mathbb{R}^{d_5}$ of object $i$ seen during training, is obtained by applying a CNN $A$ to a (ground truth) cropped image $I'_i$ of that object, resized to a fixed resolution of $64\times 64$. $d_5$ was set arbitrarily to 32 to reflect that it has less information than that of the entire object, which is embedded in $\mathbb{R}^{d_1}$.

The way in which the data flows through the sub-networks, as depicted in Fig.~\ref{fig:arch}, is captured by the equations:\\
\noindent\begin{minipage}[c]{0.46\linewidth}
\begin{align}
u_i &= G(\{n_i\},\{e_{ij}\})\label{eq:1}\\
m_i &= M(u_i,z_i)\label{eq:2}\\
b_i &= B(u_i)\label{eq:2b}
\end{align}
\end{minipage}
\hfill
\begin{minipage}[c]{0.52\linewidth}
\begin{align}
a_i &= A(I'_i) \label{eq:3}\\
t &= T(\{c_i,m_i,b_i,a_i\}) \label{eq:4}\\
p &= R(t)\label{eq:5}
\end{align}
\end{minipage}
\smallskip

where $G$ is the graph convolutional network~\cite{Johnson2018ImageGF,Scarselli2009TheGN} that generates the per-object layout embedding, $M$ and $B$ are the networks that generate the object's mask and its bounding box, respectively, $T$ is the fixed (unlearned) function that maps the various object embeddings to a tensor $t$. Finally, $R$ is the encoder-decoder network that outputs an image $p\in \mathbb{R}^{H \times W\times 3}$ based on $t$. The exact architecture of each network is provided in the appendix. 

The function $T$ constructs the tensor $t$ as a sum of per-object tensors $t_i \in \mathbb{R}^{H \times W \times (d_5+c)}$, where $c$ is the number of objects. First, the mask $m_i$ is shifted and scaled, according to the bounding box $b_i$, resulting in a mask $m^{HW}_i$ of size $H\times W$. Then, a first tensor $t_i^1$ of size $H \times W \times d_5$ is formed as the tensor product of $m^{HW}_i$ and $a_i$. Similarly, a second tensor $t_i^2 \in \mathbb{R}^{H \times W \times c}$ is formed as the tensor product of $m^{HW}_i$ and the one hot vector of length $c$ encoding class $c_i$. The tensor $t_i$ is a concatenation of the two tensors $t_i^1$ and $t_i^2$ along the third dimension.

For performing adversarial training of the appearance embedding network $A$, we create two other tensors: $t'$ and $t''$. The first one is obtained by employing the ground truth bounding box $b'_i$ of object $i$ and the ground truth segmentation mask $m'_i$. The second one is obtained by incorporating the same ground truth bounding box and mask in a counter-factual way, by replacing $a_i$ with $a_k$, where $a_k$ is an appearance embedding of an object image $I'_k$ of a different object from the same class $c_i$, i.e., $a_k = A(I'_k)$, $c_i=c_k$ and 
\begin{align}
t' &= T(\{c_i,m'_i,b'_i,a_i\}) \label{eq:6}\\
t'' &= T(\{c_i,m'_i,b'_i,a_k\}) \label{eq:7}
\end{align}

During training, in half of the training samples, the location and size information vectors $l_i$ are zeroed, in order to allow the network to generate layouts, even when this information is not available.

\subsection{Training Loss Terms}

The loss used to optimize the networks contains multiple terms, which is not surprising, given the need to train five networks (not including the adversarial discriminators mentioned below) and two vector embeddings ($o_i$ and $e_{ij})$.
\begin{multline}
L = \mathcal{L}_\text{Rec} + \lambda_1 \mathcal{L}_\text{box} + \lambda_2 \mathcal{L}_\text{perceptual} + \lambda_3 \mathcal{L}_\text{D-mask} + \lambda_4 \mathcal{L}_\text{D-image} \\+ \lambda_5 \mathcal{L}_\text{D-object} + \lambda_6 \mathcal{L}_\text{FM-mask} + \lambda_7 \mathcal{L}_\text{FM-image}
\end{multline}

where in our experiments we set $\lambda_1=\lambda_2=\lambda_6=\lambda_7=10$, $\lambda_3=\lambda_4=1$, $\lambda_5=0.1$.

The reconstruction loss $\mathcal{L}_\text{Rec}$ is the L1 difference between the reconstructed image $p$ and the ground truth training image. The box loss $\mathcal{L}_\text{box}$ is the MSE between the computed $b_i$ (summed over all objects) and the ground truth bounding box $b'_i$. Note that unlike~\cite{Johnson2018ImageGF}, we do not employ a mask loss, since our mask contains a stochastic element (Eq.~\ref{eq:2}). The perceptual loss $\mathcal{L}_\text{perceptual}= \sum_{u\in U}\frac{1}{u}||F^{u}(p)-F^{u}(p')||_1$~\cite{perceptual} compares the generated image with the ground truth training image $p'$, using the activations $F^u$ of the \textit{VGG} network~\cite{vggvergydeep} at layer $u$ in a set of predefined layers $U$. 

Our method employs three discriminators $D_{\text{mask}}$, $D_\text{object}$, and $D_\text{image}$. The mask discriminator employs a Least Squares GAN (\textit{LS-GAN}~\cite{lsgan}) and is conditioned on the object's class $c_i$. Recall that $m'_i$ is the real mask of object $i$ and $m_i$, the generated mask, which depends on a random variable $z_i$. The GAN loss associated with the mask discriminator is given by
\begin{multline}
\mathcal{L}_{D-mask} = [\log D_\text{mask}(m'_i, c_i)] + \\ \mathop{\mathbb{E}}_{z\sim\mathcal{N}(0,1)^{64}}[log(1-D_\text{mask}(M(u_i, z), c_i)]
\end{multline}
For the purpose of training $D_\text{mask}$, we minimize $-\mathcal{L}_{D-mask}$. The second discriminator $D_{image}$, is used for training in an adversarial manner three networks $R$, $M$, and $A$. The loss of $\mathcal{L}_\text{D-image}$ is a compound loss that is given as
\begin{equation}
\mathcal{L}_\text{D-image} = \mathcal{L}_\text{real} - \mathcal{L}_\text{fake-image} - \mathcal{L}_\text{fake-layout} + \mathcal{L}_\text{alt-appearance}\nonumber
\end{equation}
where 
\begin{align}
    \mathcal{L}_\text{real} &= \log D_\text{image}(t', p')\\
    \mathcal{L}_\text{fake-image} &= \log(1- D_\text{image}(t', p))\\
    \mathcal{L}_\text{fake-layout} &= \log(1- D_\text{image}(t, p'))\\
    \mathcal{L}_\text{alt-appearance} &= \log(1- D_\text{image}(t'', p'))
\end{align}

The goal of the compound loss is to make sure that the generated image $p$, given a ground truth layout tensor $t'$ is indistinguishable from the real image $p'$, and that this is true, even if the layout tensor $t$ is based on estimated bounding boxes and masks (unlike $t'$). In addition, we would like the ground truth image to be a poor match for a counterfactual appearance vector, as given in $t''$.

Following \cite{wang2018pix2pixHD} we use a multi-scale \textit{LS-GAN} with two scales. In other words, $\mathcal{L}_\text{D-image}$ is computed at the full scale and at half scale (using two different discriminators), and both terms are summed up to obtain the actual $\mathcal{L}_\text{D-image}$.

The third discriminator, $D_\text{object}$, guarantees that the generated objects, one by one, look real. For this purpose, we crop $p$ using the bounding boxes $b_i$ to create object images $I_i$. Recall that $I'_i$ are ground truth crops of images, obtained from the ground truth image $p'$, using the ground truth bounding boxes $b'$. 
\begin{equation}
\mathcal{L}_\text{D-object} = \sum_{i=1}^n \log D_\text{object}(I'_i)-\log D_\text{object}(I_i)
\end{equation}
$D_\text{object}$ maximizes this loss during training.

The mask feature matching loss $\mathcal{L}_\text{FM-mask}$ and the image feature matching loss $\mathcal{L}_\text{FM-image}$ are similar to the perceptual loss, i.e., they are based on the L1 difference in the activation. However, instead of the \textit{VGG} loss, the discriminators are used, as in~\cite{Salimans2016ImprovedTF}. In these losses, all layers are used. $\mathcal{L}_\text{FM-mask}$ compares the activations of the generated mask $m_i$ and the real mask $m'_i$ (the discriminator $D_\text{mask}$ also takes the class $c_i$ as input). The other feature matching loss $\mathcal{L}_\text{FM-image}$ compares the activations of $D_\text{image}(t,p)$ with those of the ground truth layout tensor and image $D_\text{image}(t',p')$.

\newcommand{\photo}[1]{%
    \includegraphics[width=0.14\linewidth, width=0.14\linewidth]{#1}
}

\begin{figure*}
\centering
\setlength\tabcolsep{0pt}%
\begin{tabularx}{\textwidth}{ccccccc}
    \photo{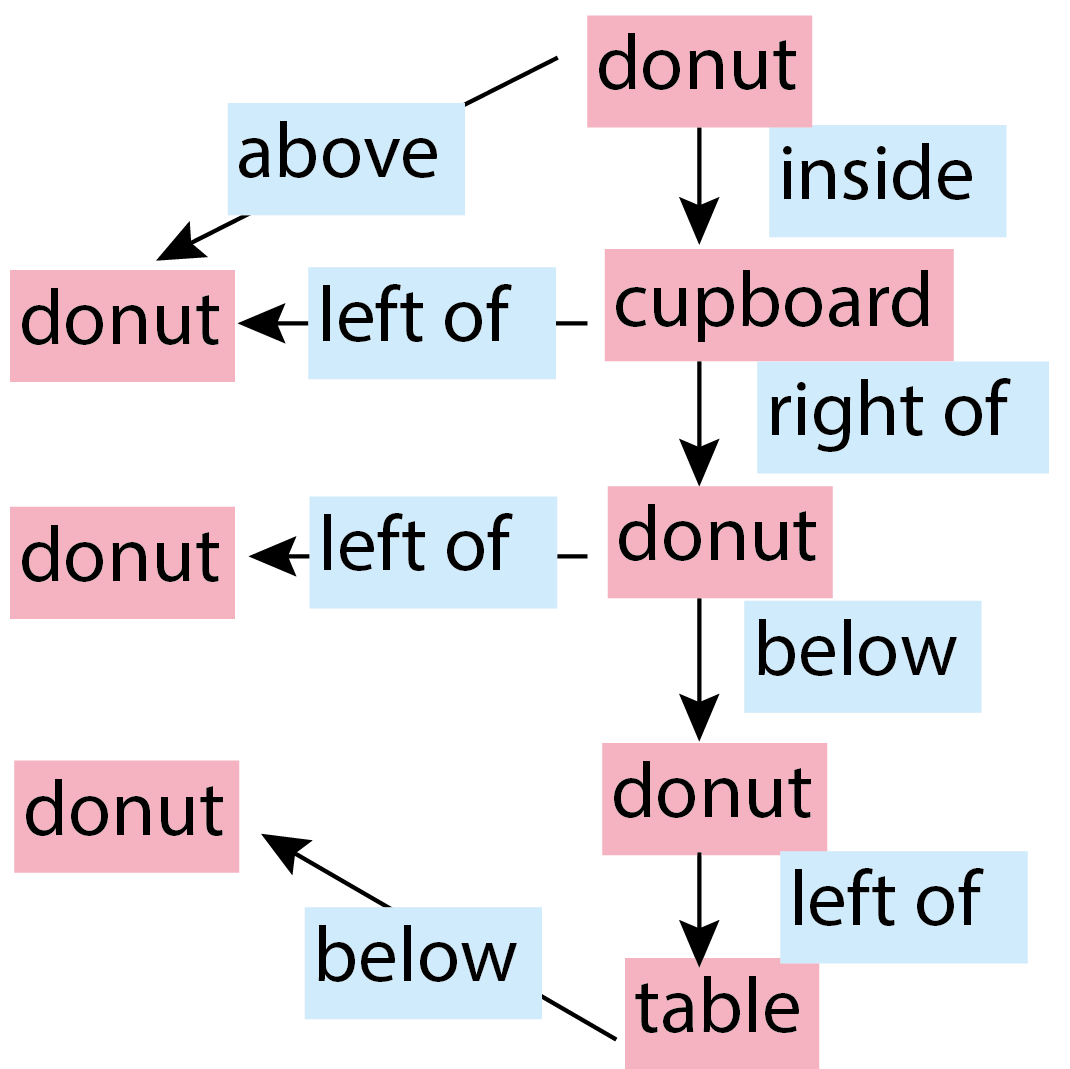} &
    \photo{} &
    \photo{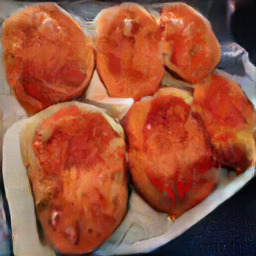} &
    \photo{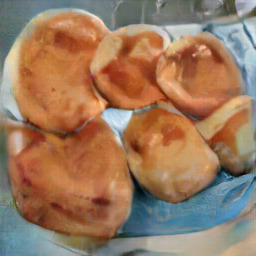} &
    \photo{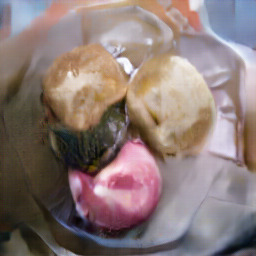} &
    \photo{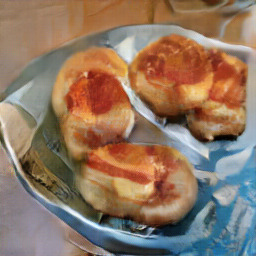} &
    \photo{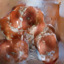} \\
    \photo{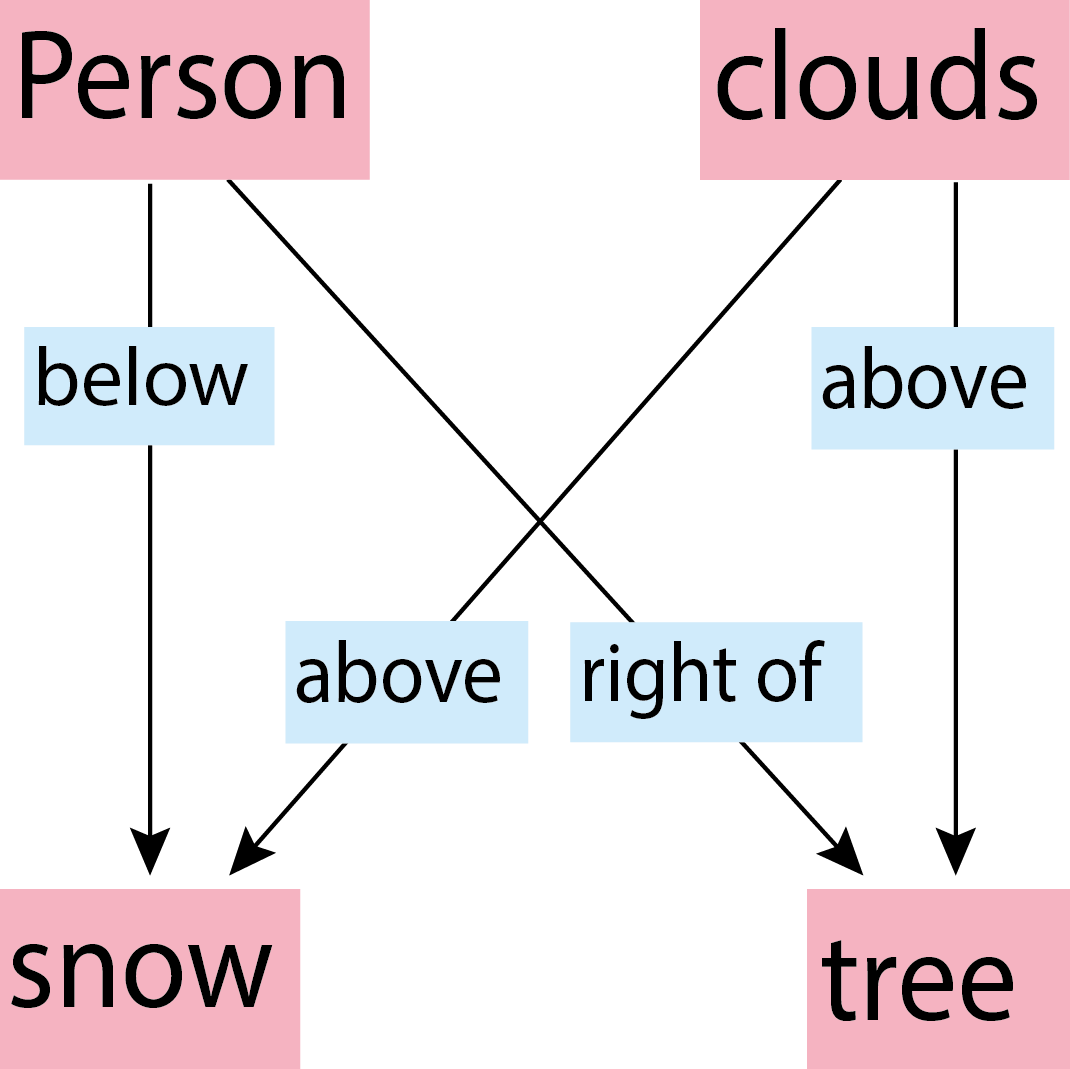} &
    \photo{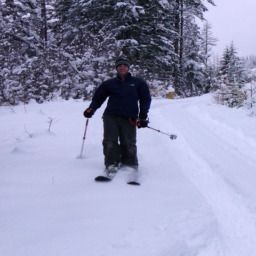} &
    \photo{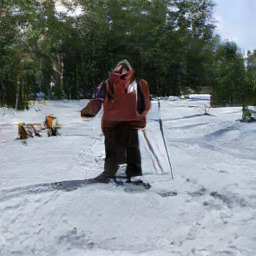} &
    \photo{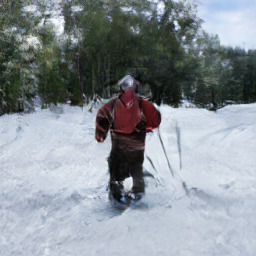} &
    \photo{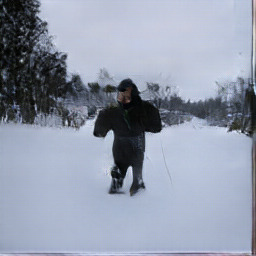} &
    \photo{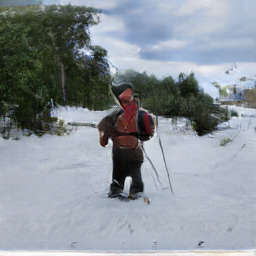} &
    \photo{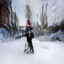} \\
    \photo{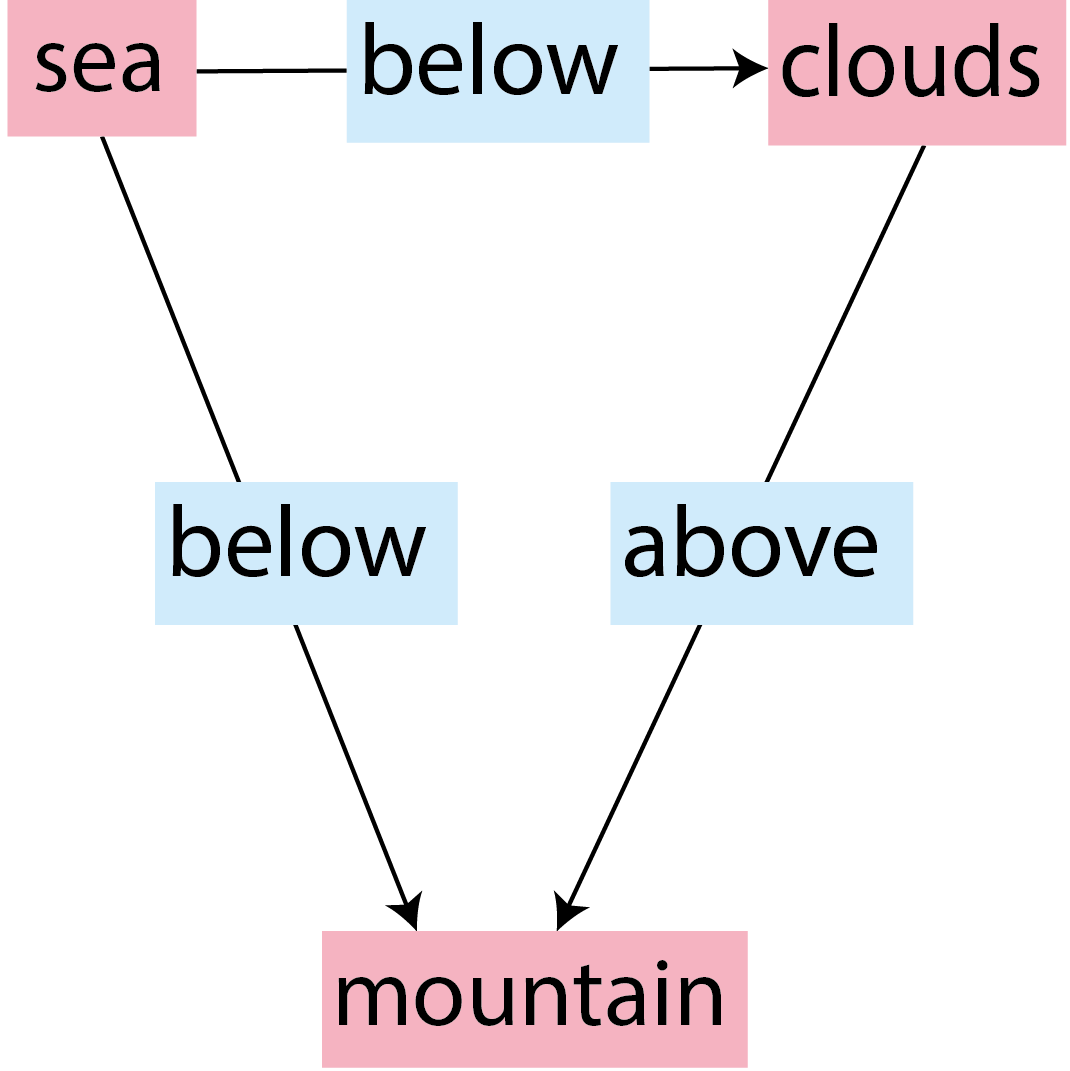} &
    \photo{} &
    \photo{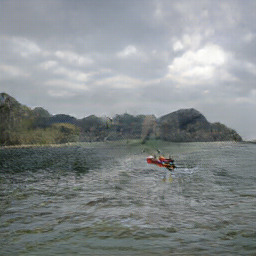} &
    \photo{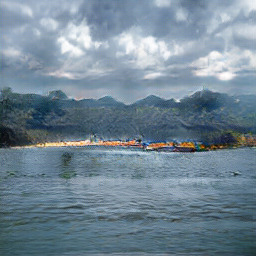} &
    \photo{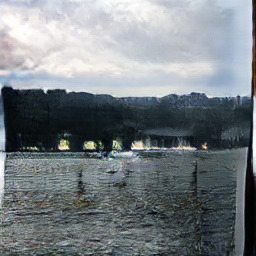} &
    \photo{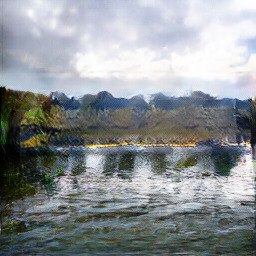} &
    \photo{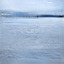} \\
    \photo{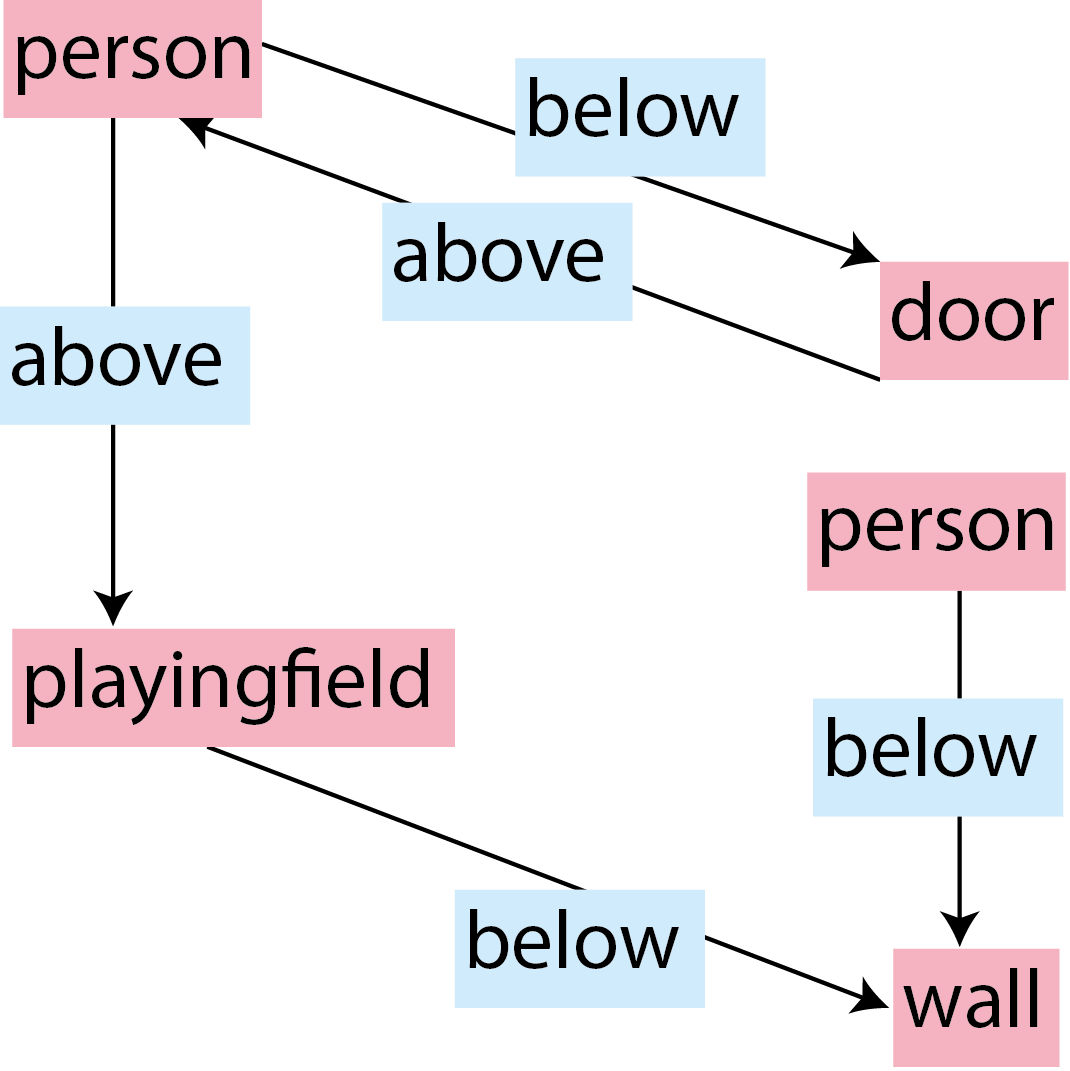} &
    \photo{} &
    \photo{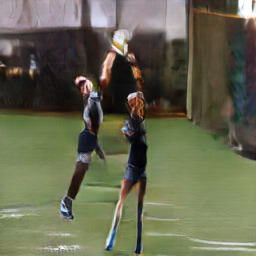} &
    \photo{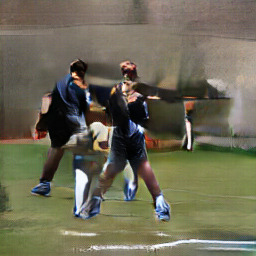} &
    \photo{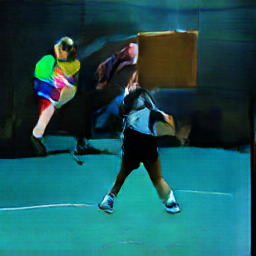} &
    \photo{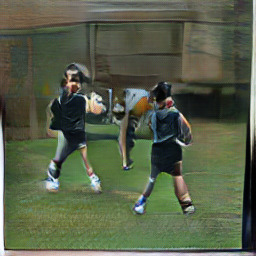} &
    \photo{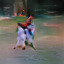} \\
    \photo{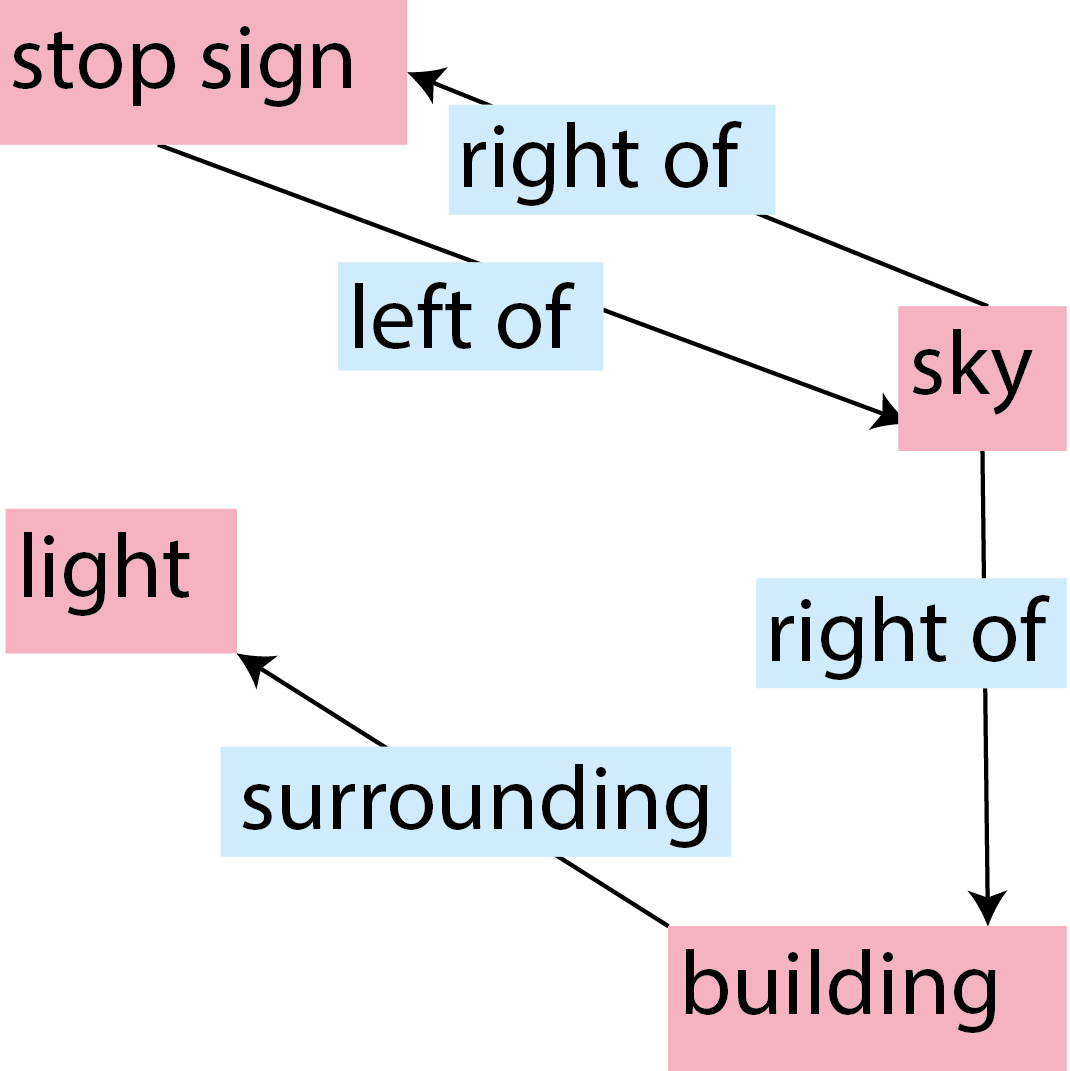} &
    \photo{} &
    \photo{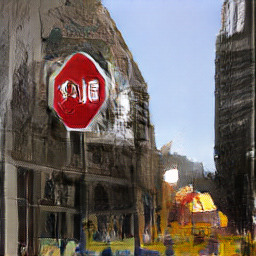} &
    \photo{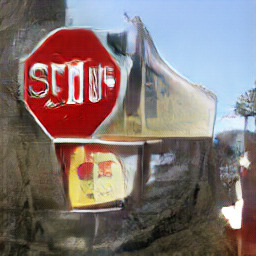} &
    \photo{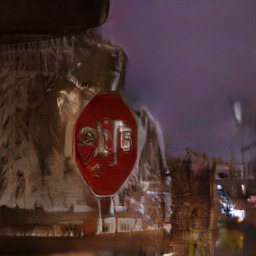} &
    \photo{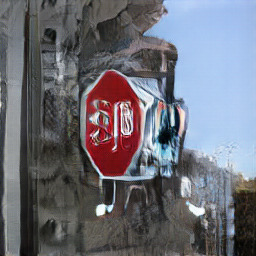} &
    \photo{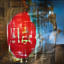} \\
    \photo{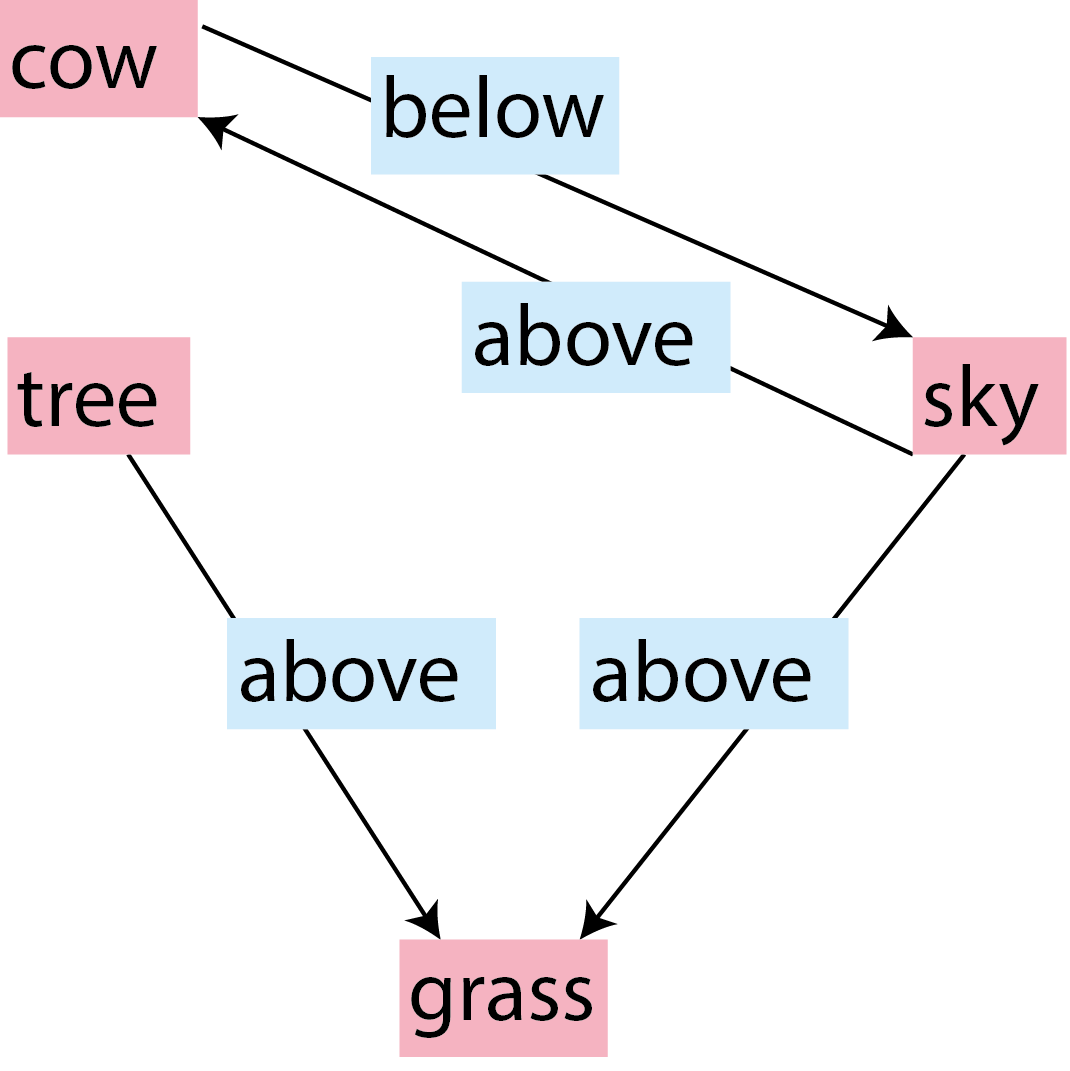} &
    \photo{} &
    \photo{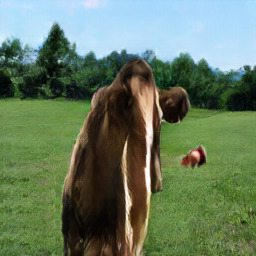} &
    \photo{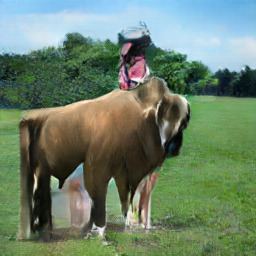} &
    \photo{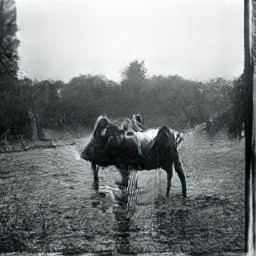} &
    \photo{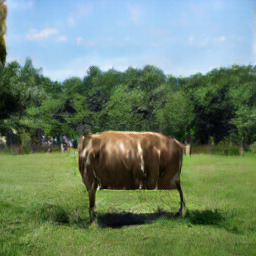} &
    \photo{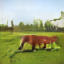} \\
    \photo{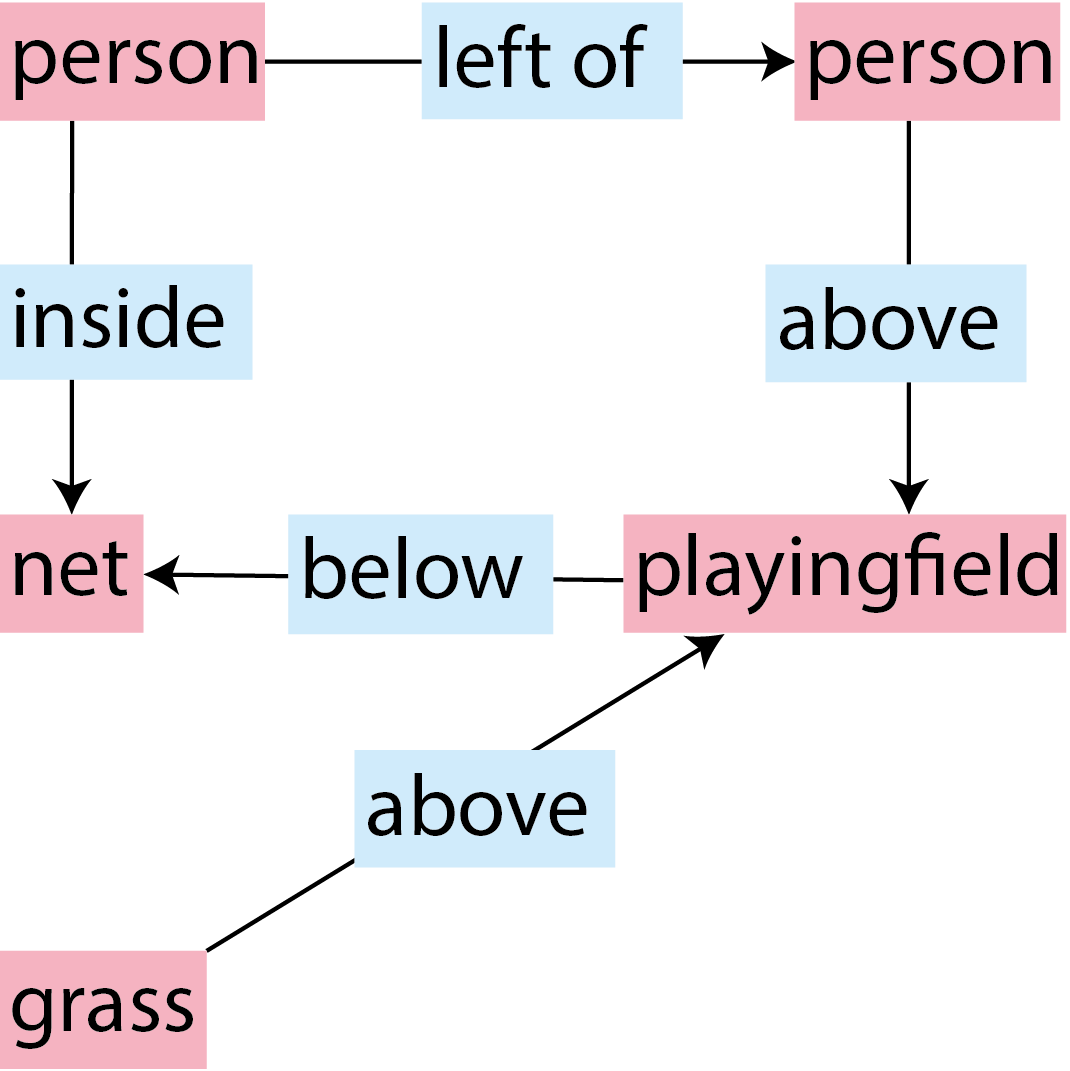} &
    \photo{} &
    \photo{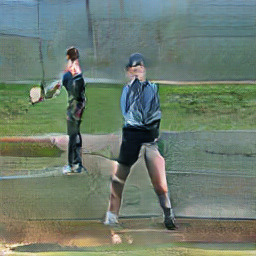} &
    \photo{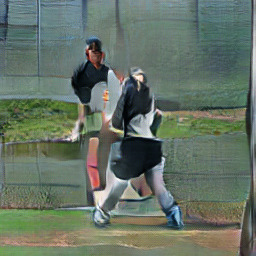} &
    \photo{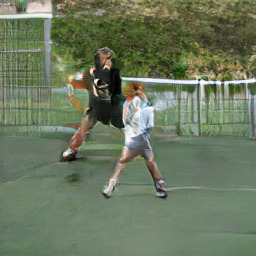} &
    \photo{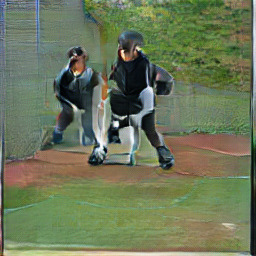} &
    \photo{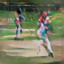} \\
(a) & (b) & (c) & (d) & (e) & (f) & (g) \\
\end{tabularx}
\caption{Image generation based on a given scene graph. Each row is a different example. (a) the scene graph, (b) the ground truth image, from which the layout was extracted, (c) our results when we used the ground truth layout of the image, similar to~\cite{Zhao2018ImageGF}, (d)  our method's results, where the appearance attributes present a random archetype and the location attributes coarsely describe the ground truth bounding box, (e) our results when we use the ground truth image to generate the appearance attributes, and the location attributes are zeroed $l_i=0$, (f) our results where $l_i=0$, and the appearance attributes are sampled from the archetypes, and (g) the results of~\cite{Johnson2018ImageGF}.} \label{fig:comparison}
\end{figure*}

\begin{figure*}
\centering
\setlength\tabcolsep{0pt}%
\begin{tabularx}{\textwidth}{ccccccc}
    \photo{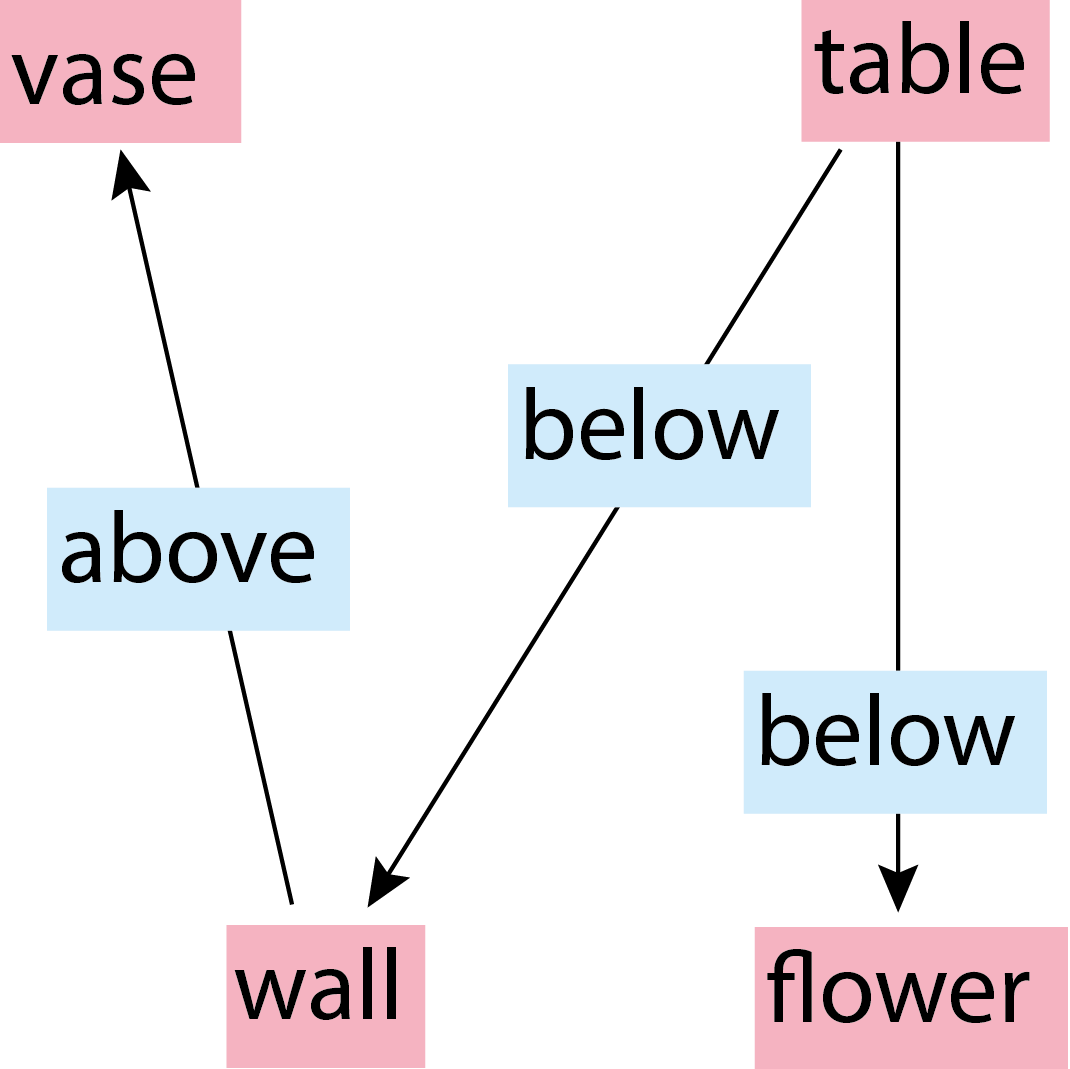} &
    \photo{} &
    \photo{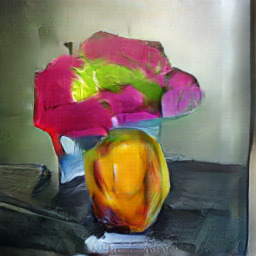} &
    \photo{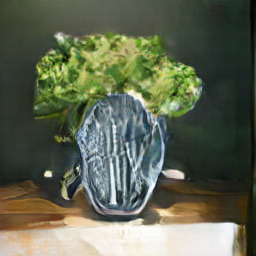} &
    \photo{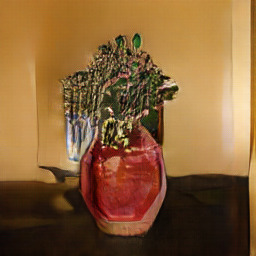} &
    \photo{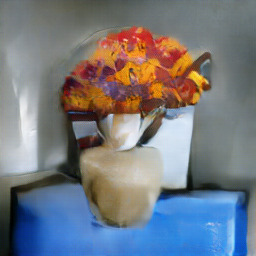} &
    \photo{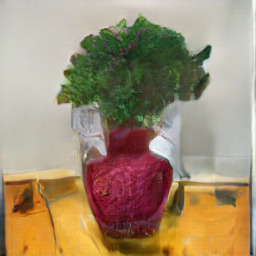} \\
    \photo{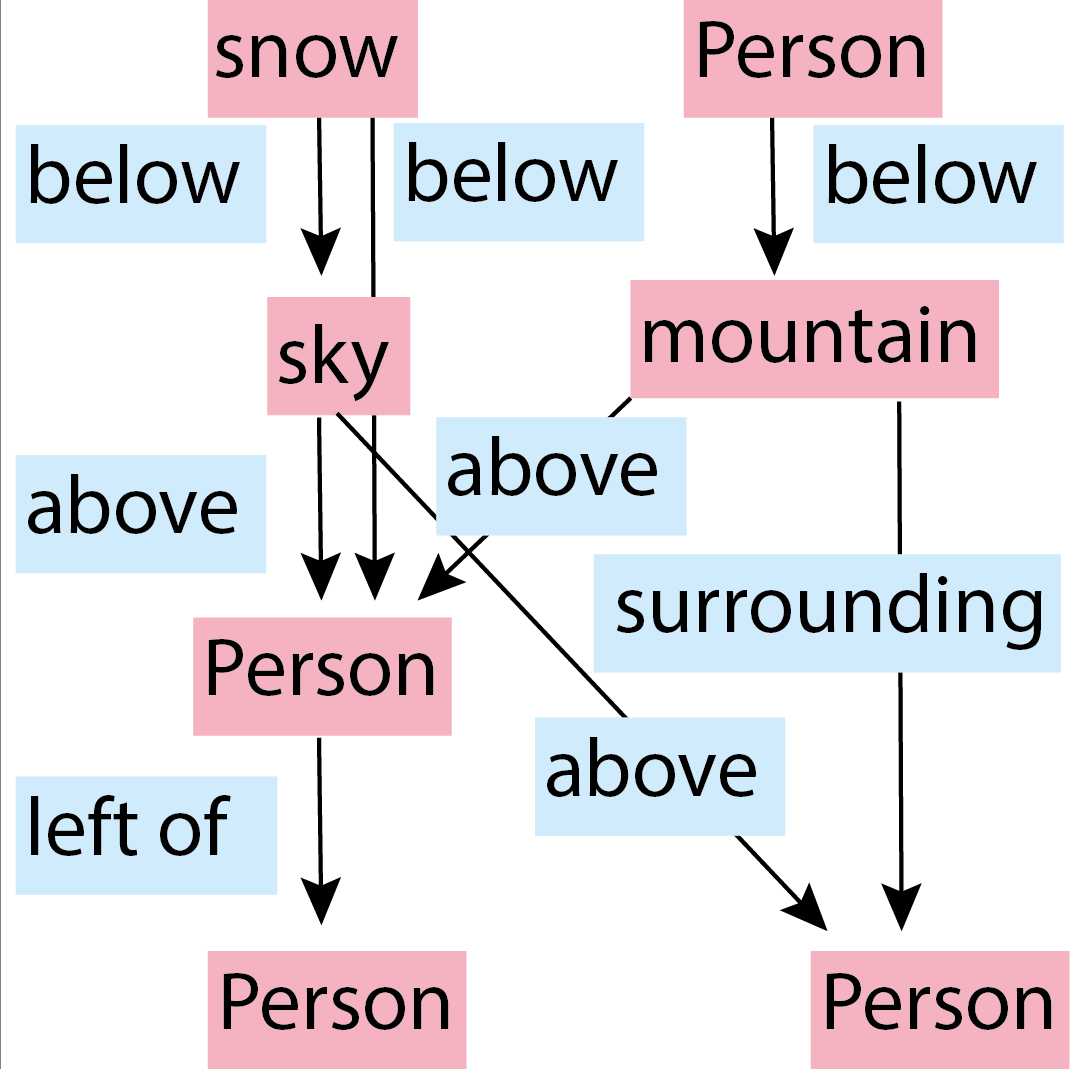} &
    \photo{} &
    \photo{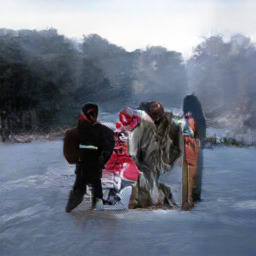} &
    \photo{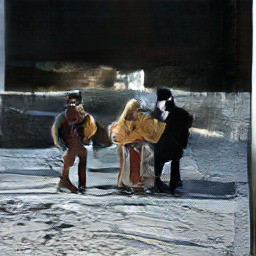} &
    \photo{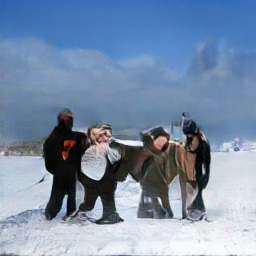} &
    \photo{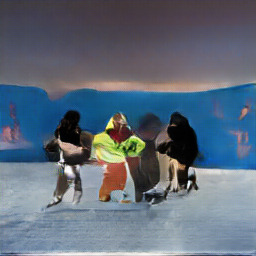} &
    \photo{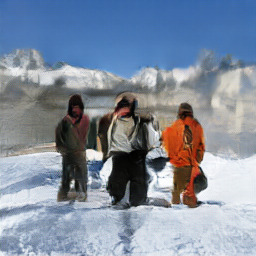} \\
    \photo{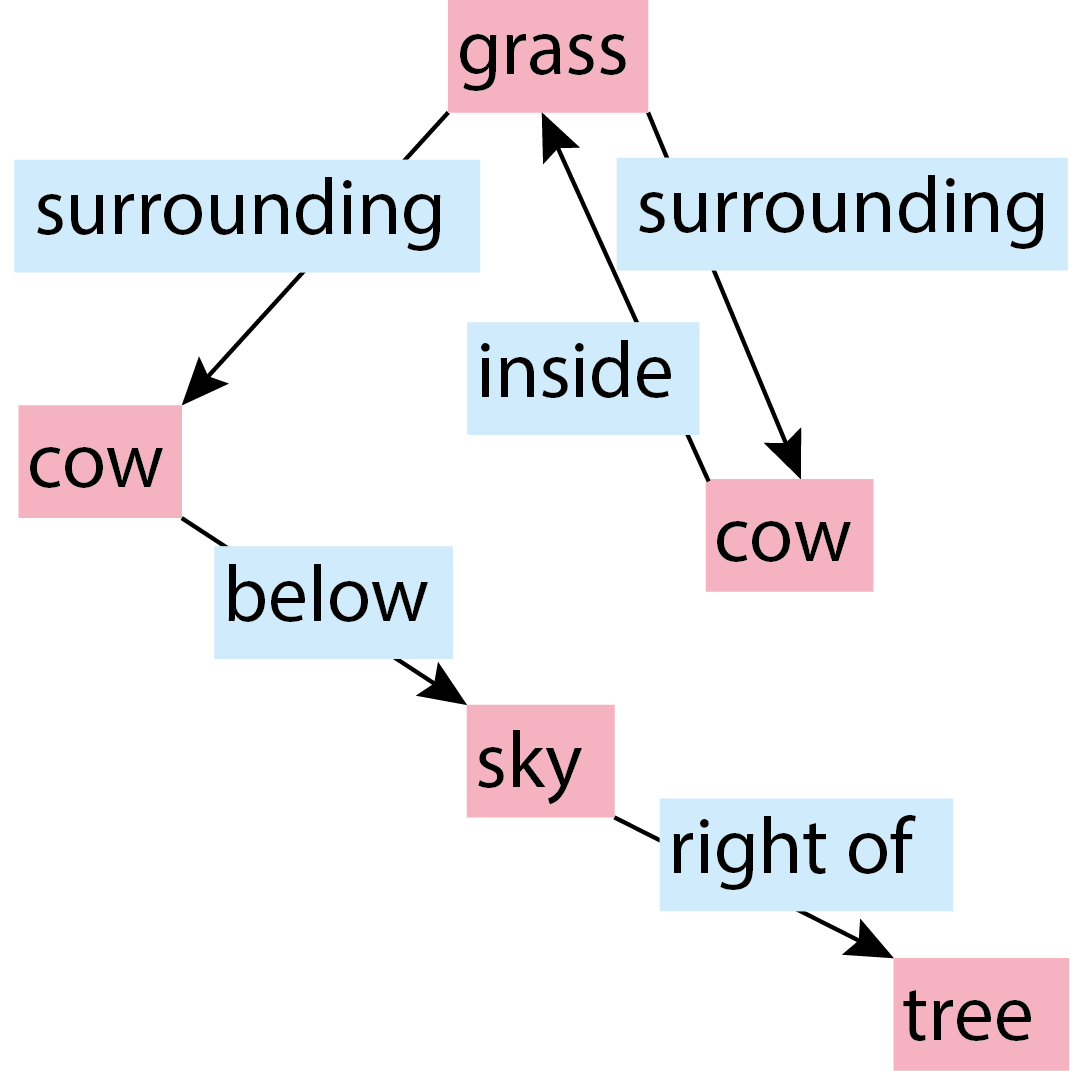} &
    \photo{} &
    \photo{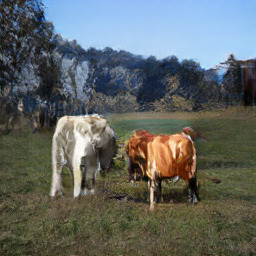} &
    \photo{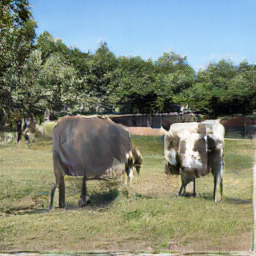} & 
    \photo{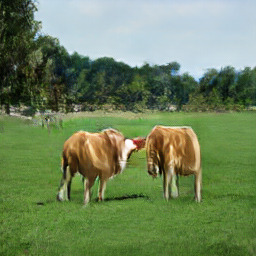} &
    \photo{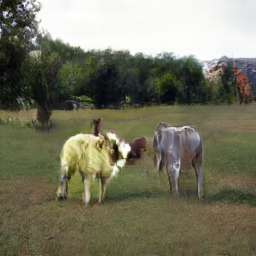} &
    \photo{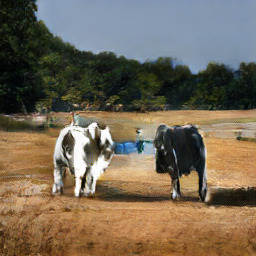} \\
    \photo{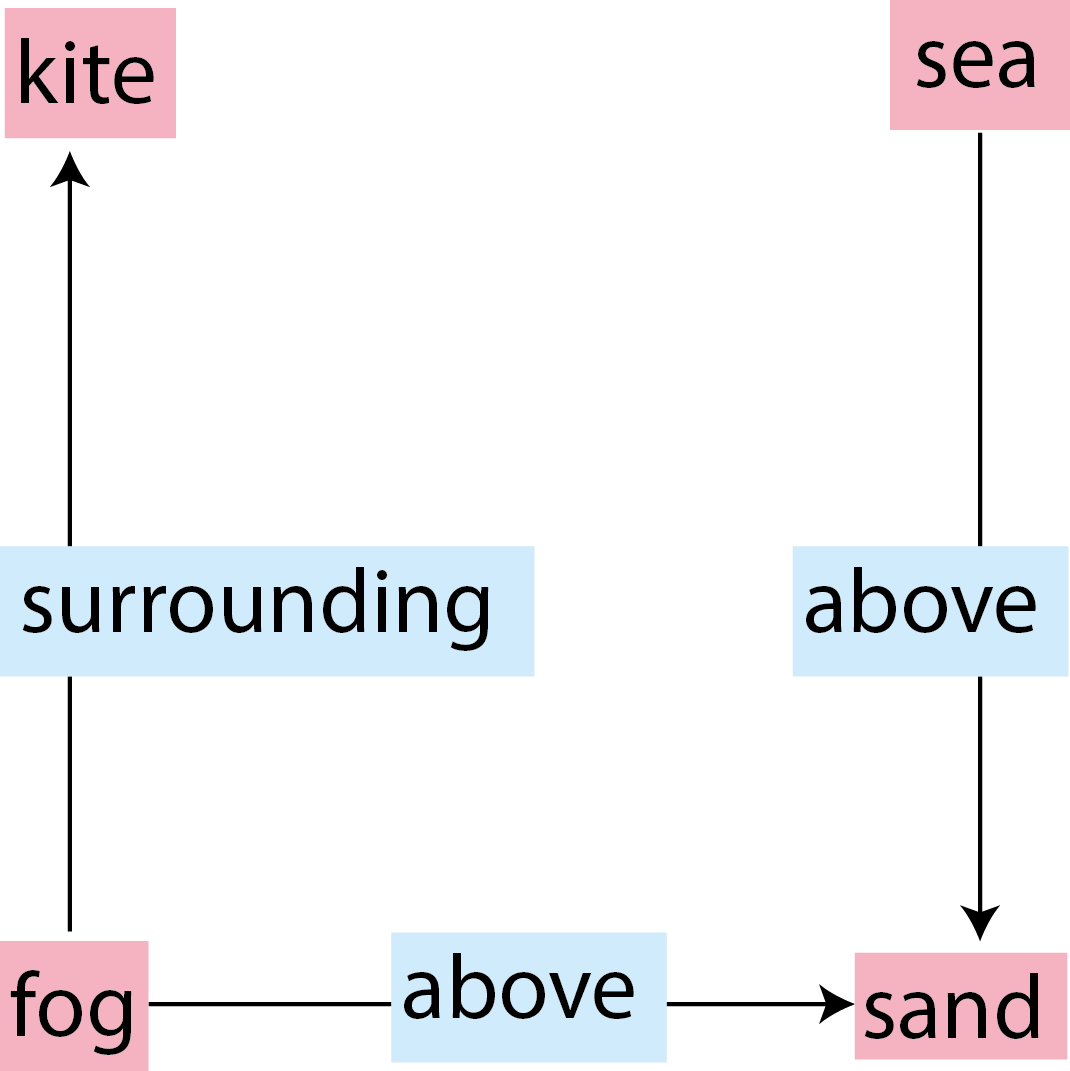} &
    \photo{} &
    \photo{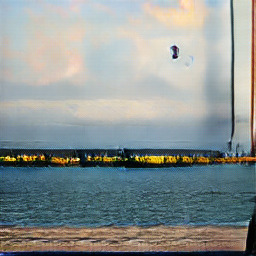} &
    \photo{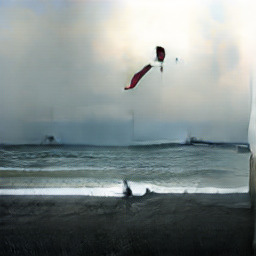} &
    \photo{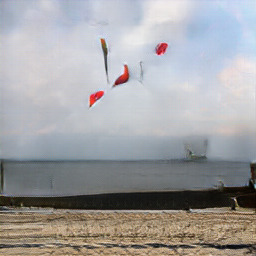} &
    \photo{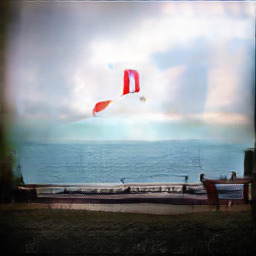} &
    \photo{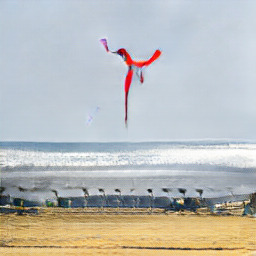} \\
    \photo{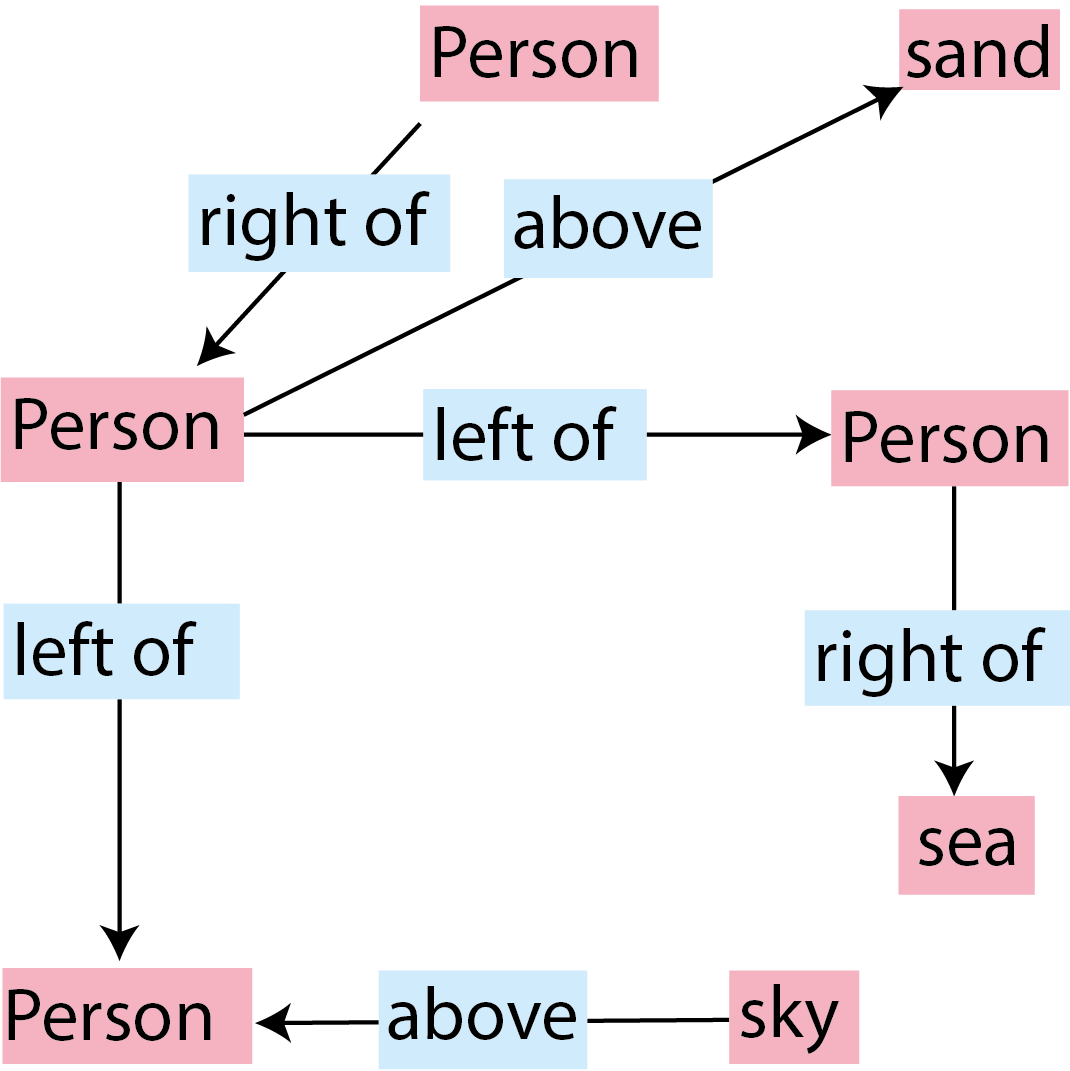} &
    \photo{} &
    \photo{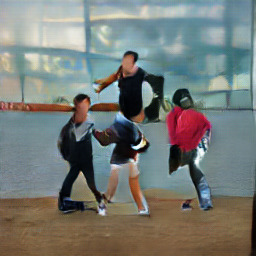} &
    \photo{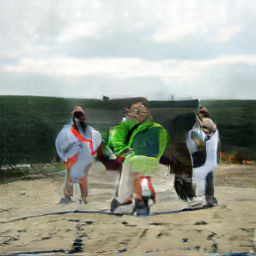} &
    \photo{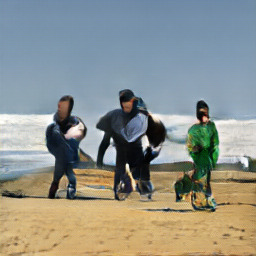} &
    \photo{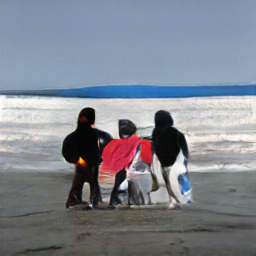} &
    \photo{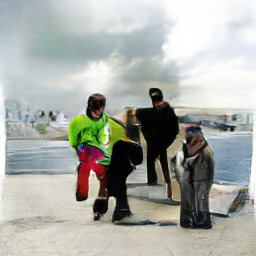}\\
    \photo{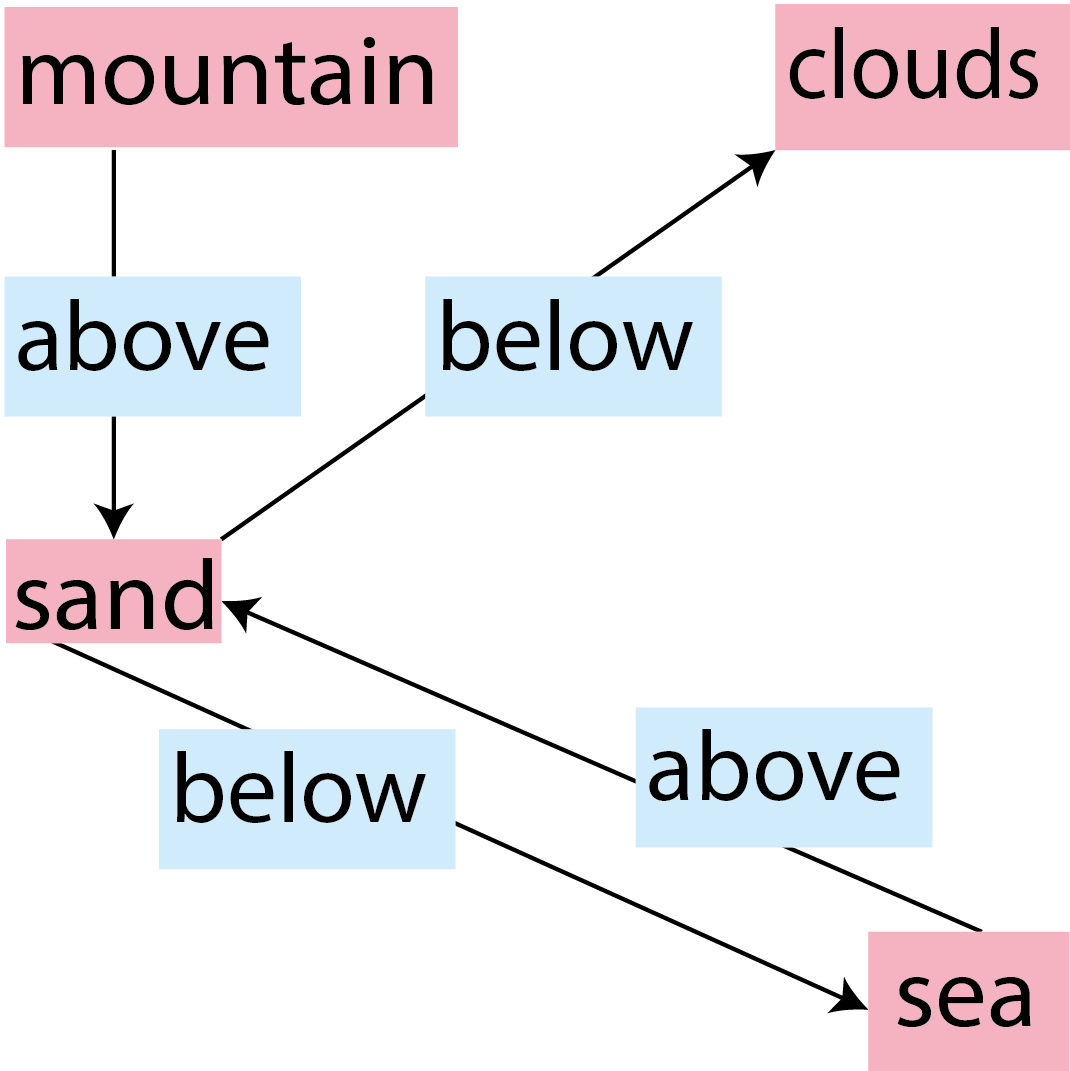} &
    \photo{} &
    \photo{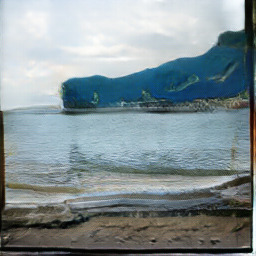} &
    \photo{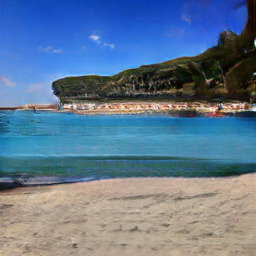} &
    \photo{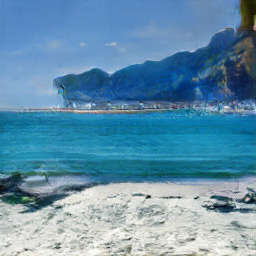} &
    \photo{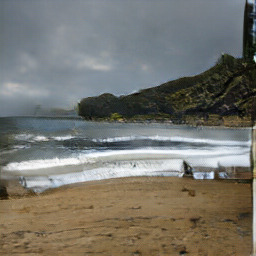} &
    \photo{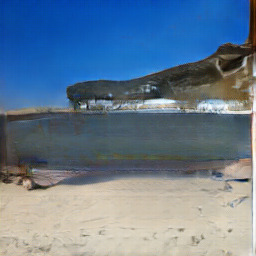} \\
(a) & (b) & (c) & (d) & (e) & (f) & (g)\\
\end{tabularx}
\caption{The diversity obtained when keeping the location attributes $l_i$ fixed at zero and sampling different appearance archetypes. (a) the scene graph, (b) the ground truth image, from which the layout was extracted, (c--g) generated images.} \label{fig:adiversity}
\end{figure*}

\begin{figure*}
\centering
\setlength\tabcolsep{0pt}%
\begin{tabularx}{\textwidth}{ccccccc}
    \photo{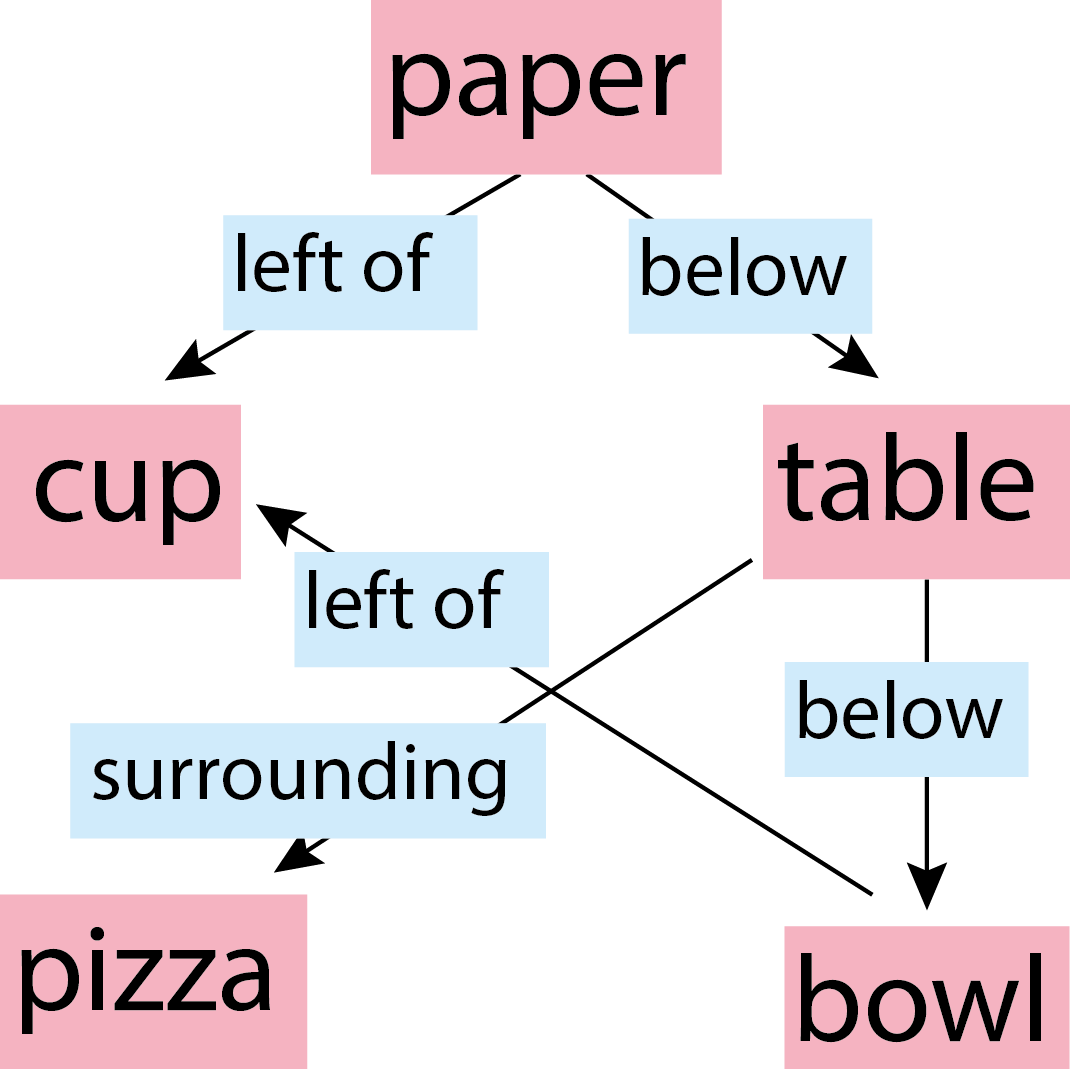} &
    \photo{} &
    \photo{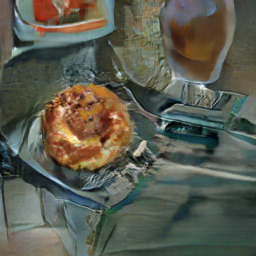} &
    \photo{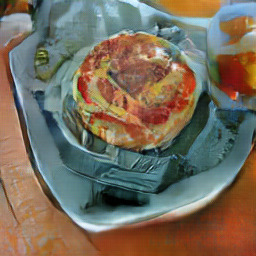} &
    \photo{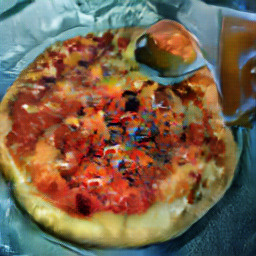} &
    \photo{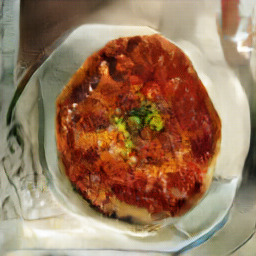} &
    \photo{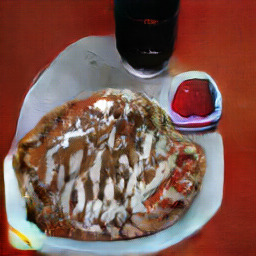} \\
    \photo{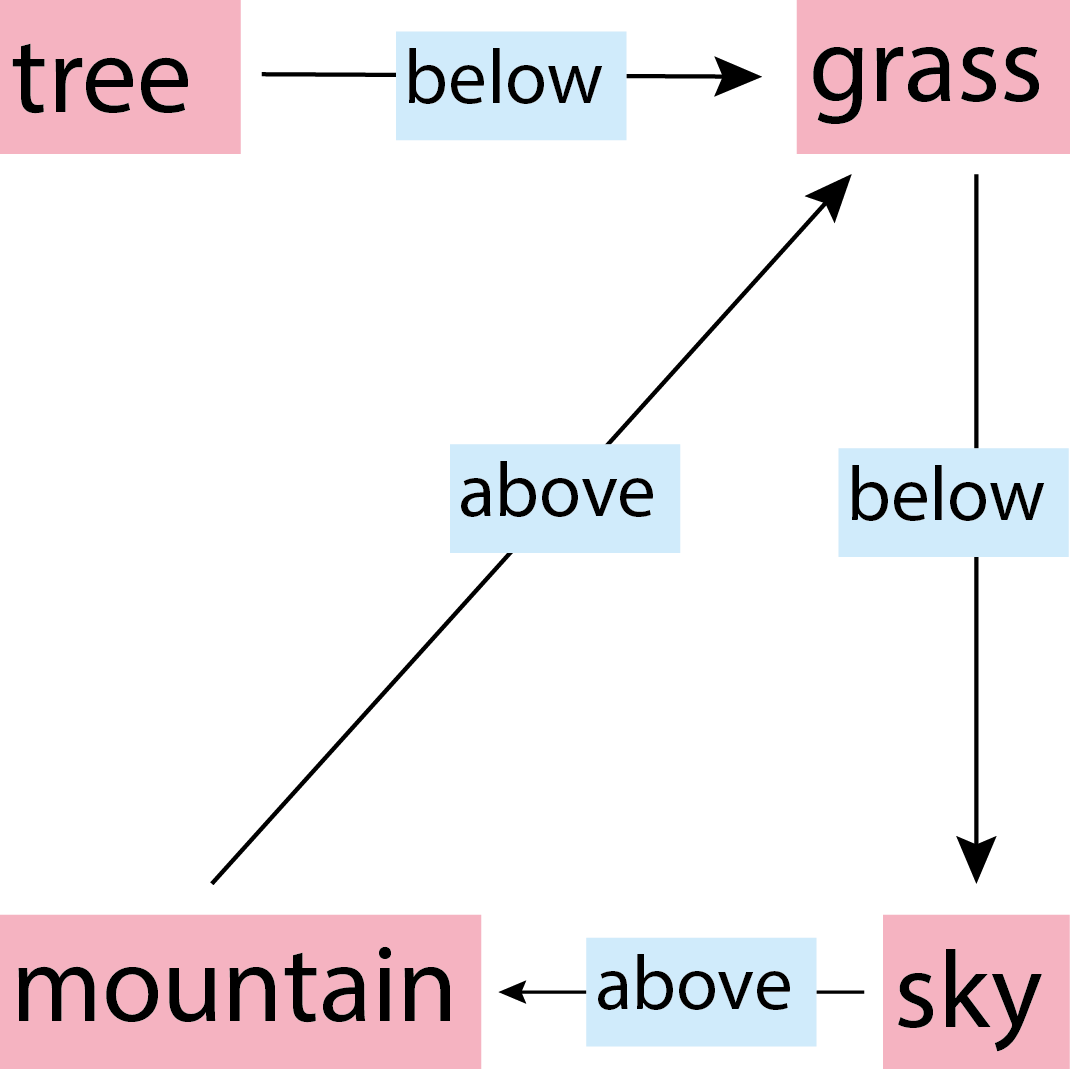} &
    \photo{} &
    \photo{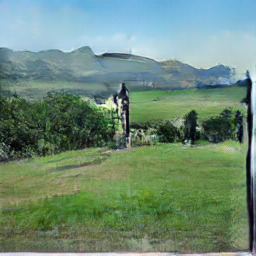} &
    \photo{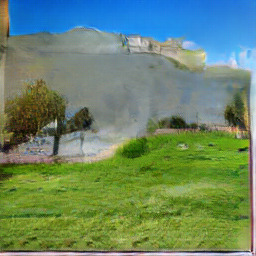} &
    \photo{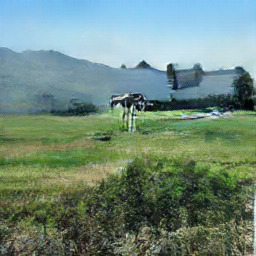} &
    \photo{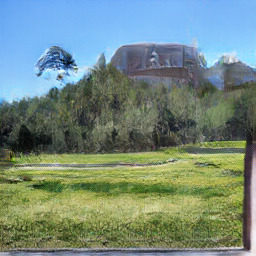} &
    \photo{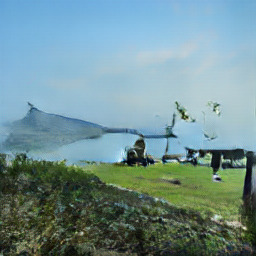} \\
    \photo{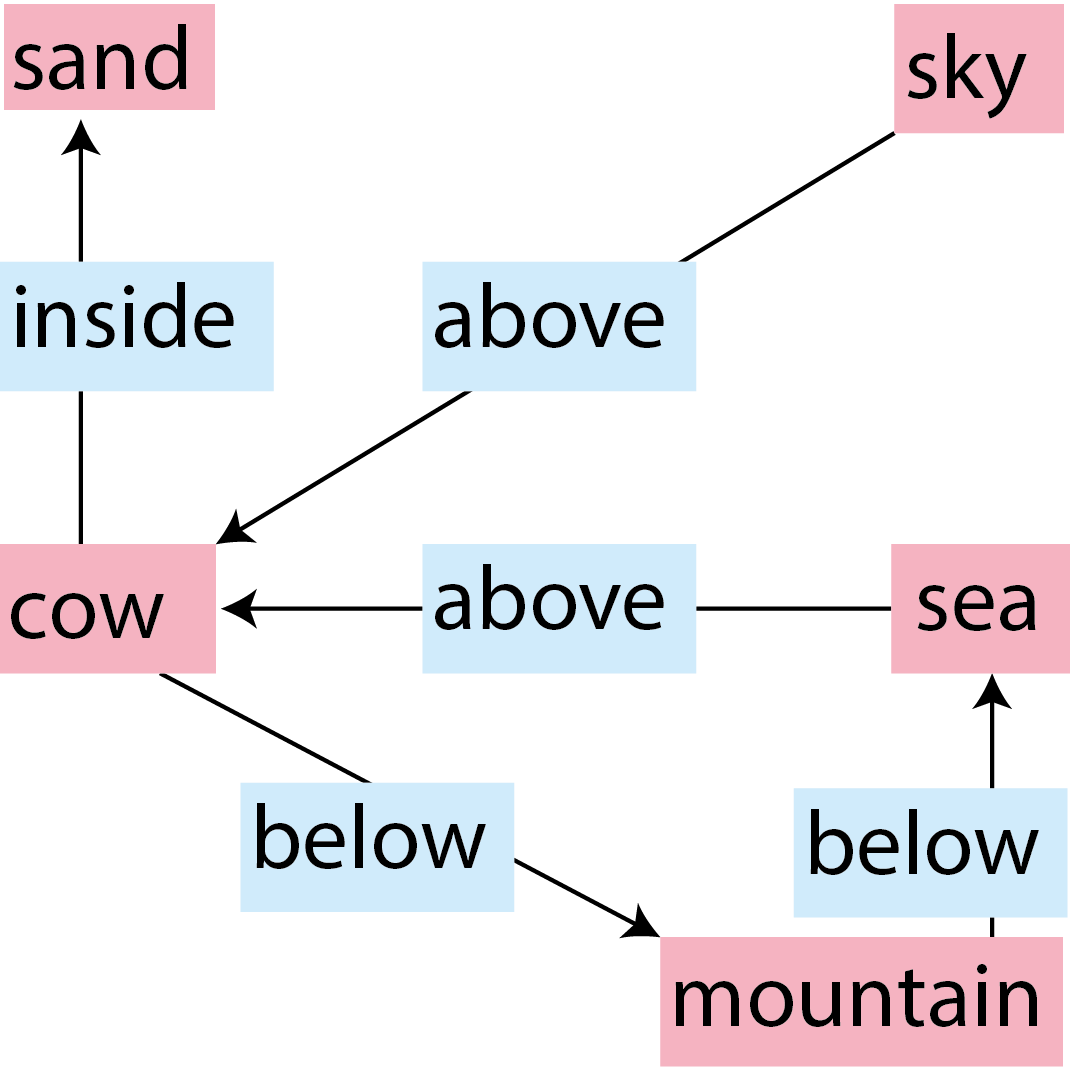} &
    \photo{} &
    \photo{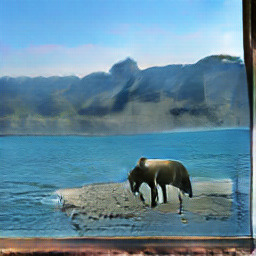} &
    \photo{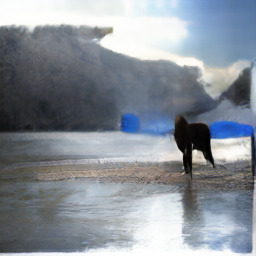} &
    \photo{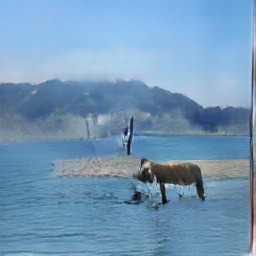} &
    \photo{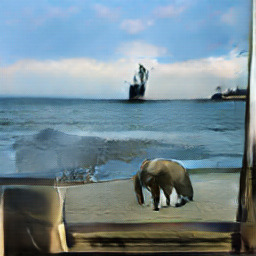} &
    \photo{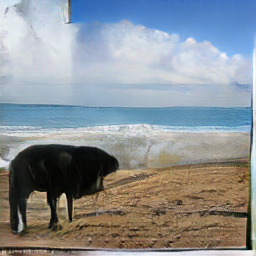} \\
    \photo{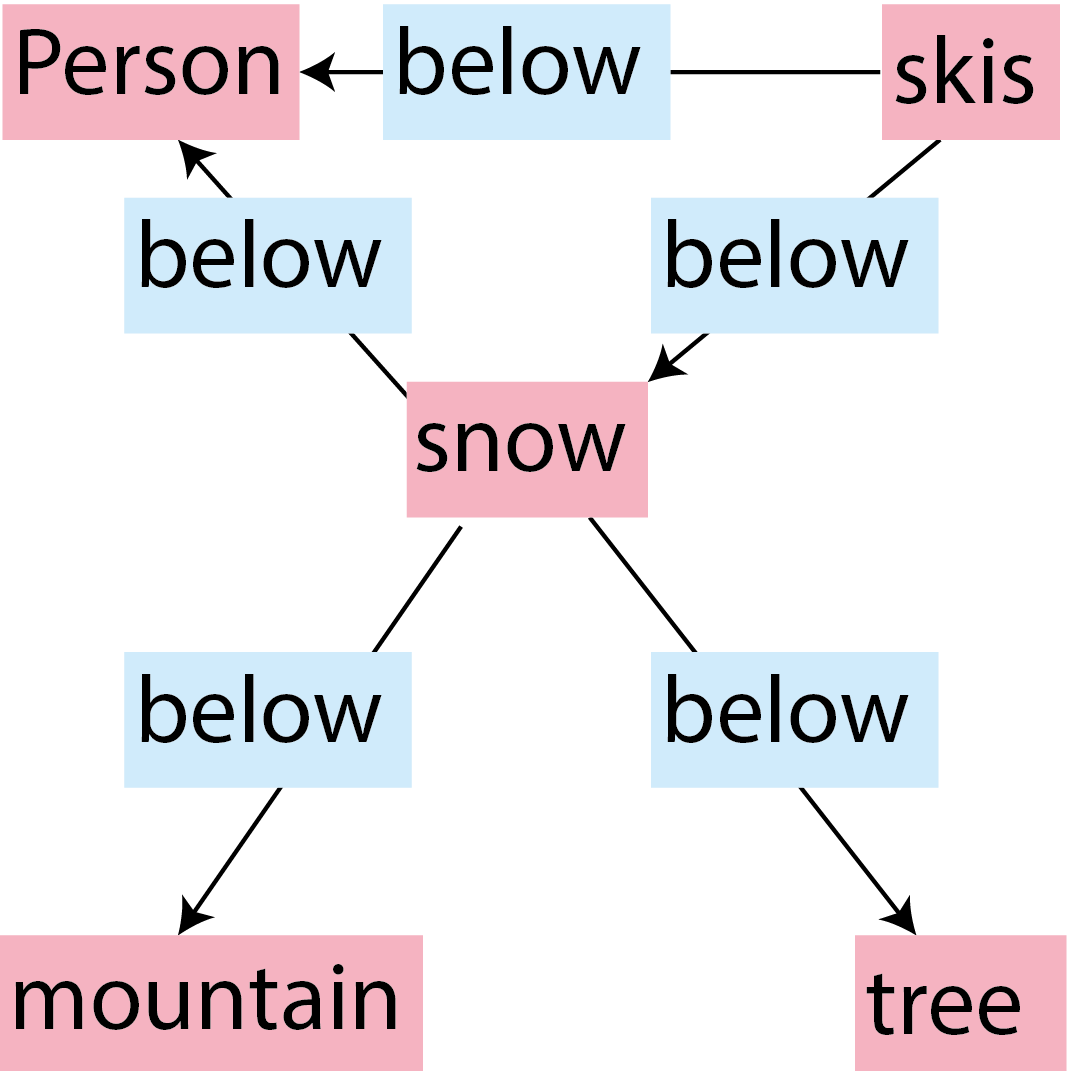} &
    \photo{} &
    \photo{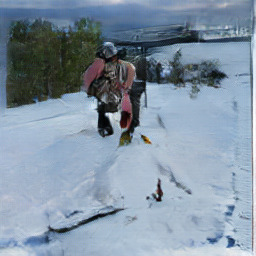} &
    \photo{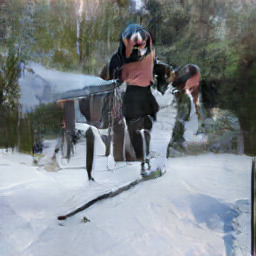} &
    \photo{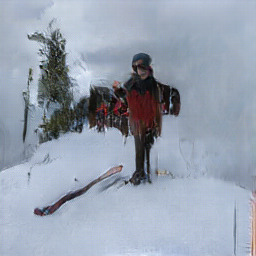} &
    \photo{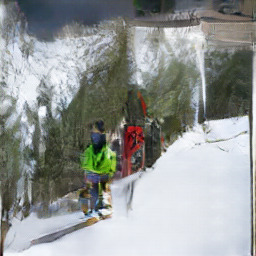} &
    \photo{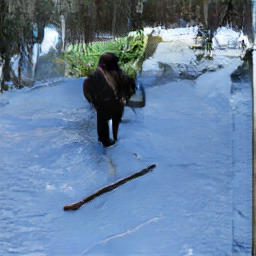} \\
    \photo{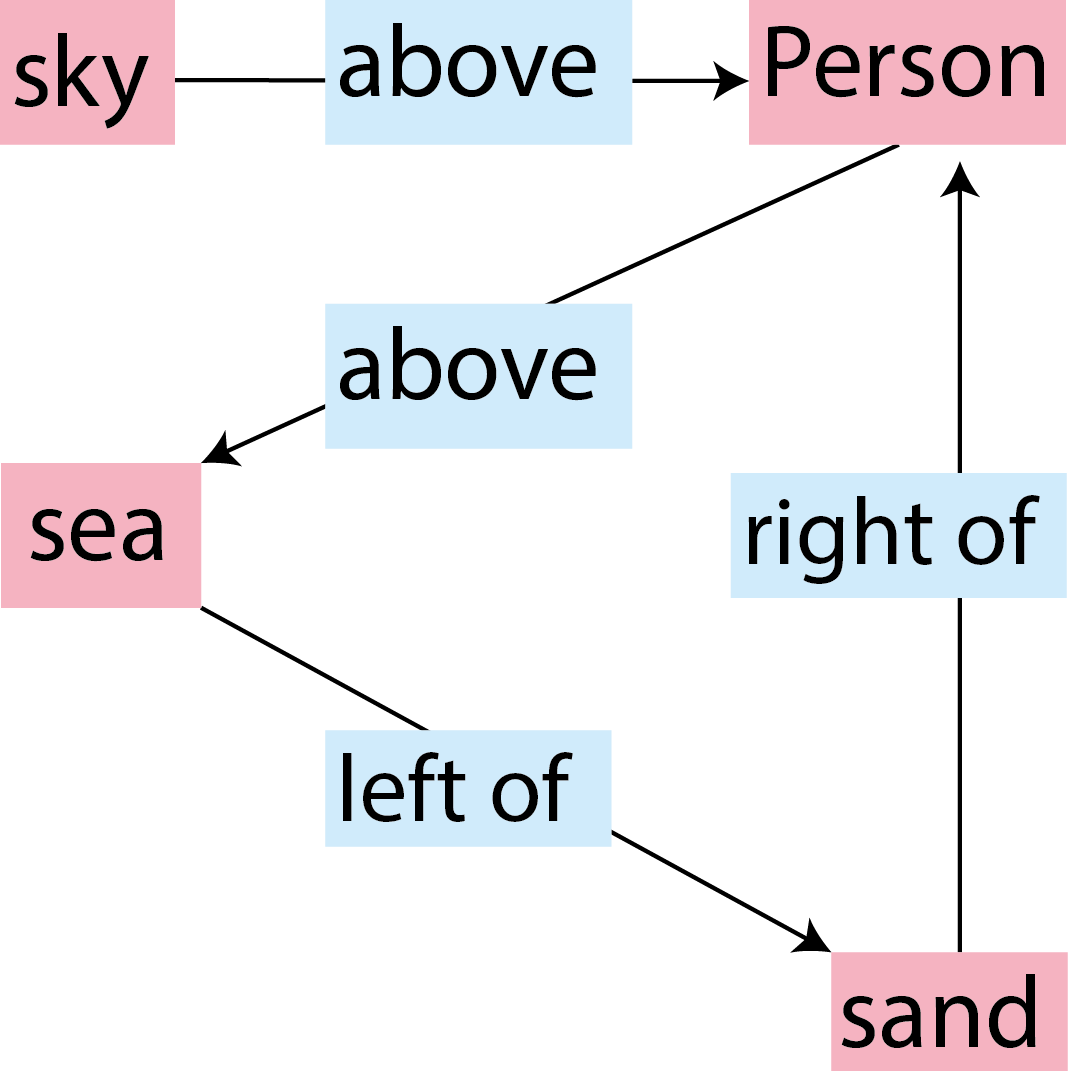} &
    \photo{} &
    \photo{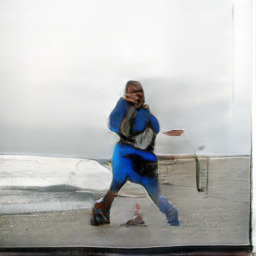} &
    \photo{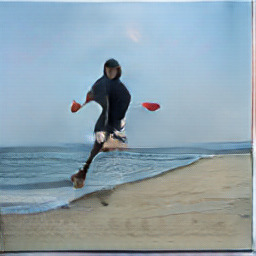} &
    \photo{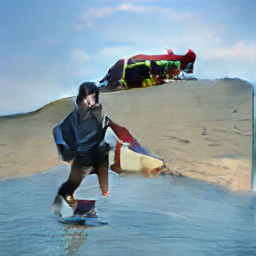} &
    \photo{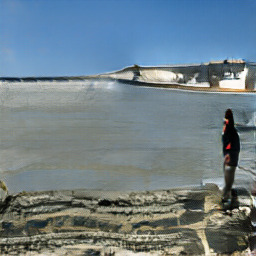} &
    \photo{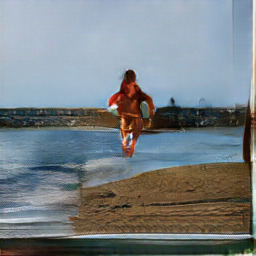} \\
(a) & (b) & (c) & (d) & (e) & (f) & (g) \\
\end{tabularx}
\caption{The diversity obtained when keeping the appearance vectors fixed and sampling from the location distribution. (a) the scene graph, (b) the ground truth image from which the layout was extracted, (c--g) generated images.} \label{fig:ldiversity}
\end{figure*}

\begin{figure*}
\centering
  \includegraphics[
  width=1\linewidth]{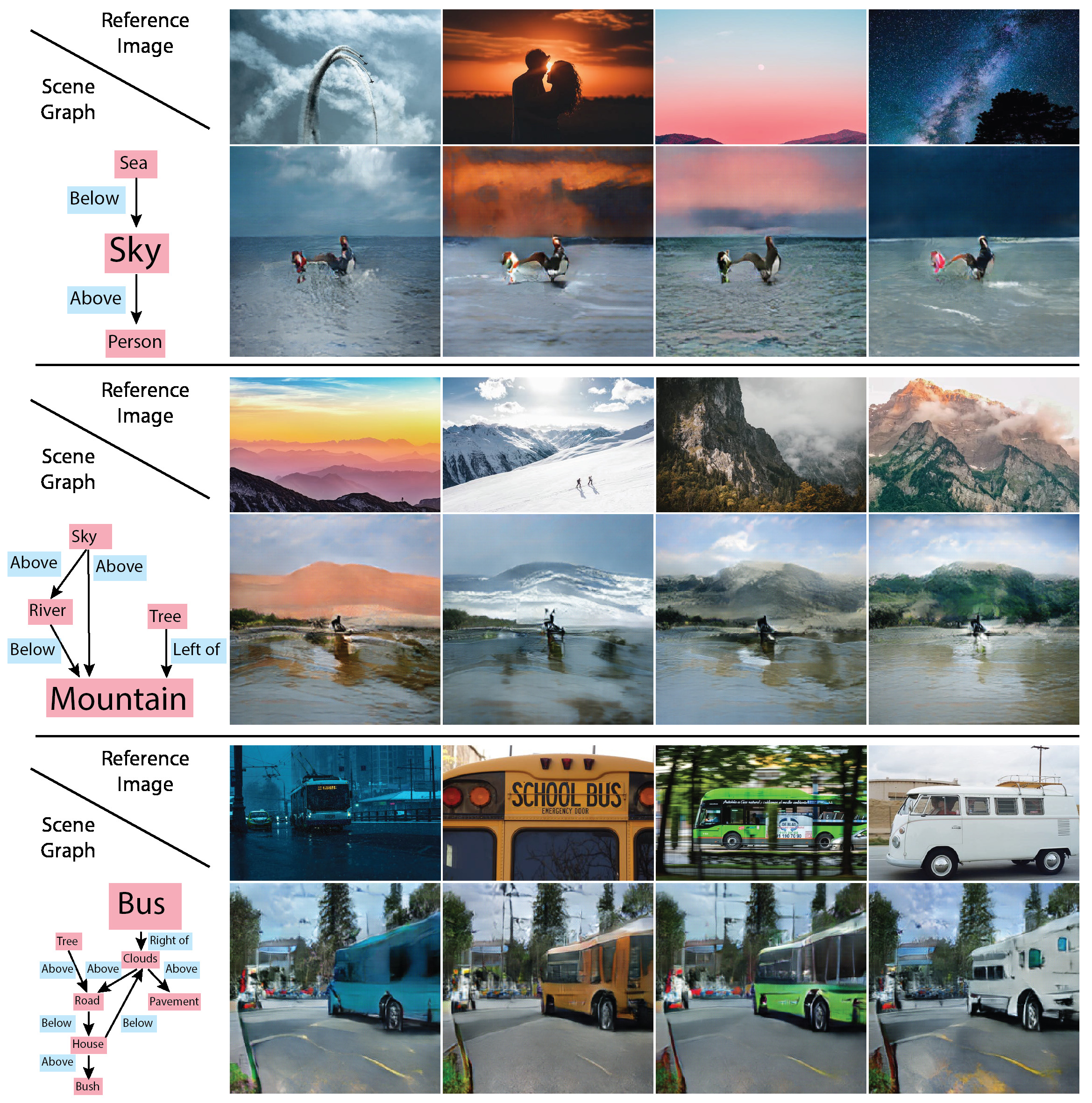}
  \caption{Duplicating an object's appearance in the generated image. Images are created based on the scene graph, such that the appearance is taken from one of five unrelated images. In this example, the sky's appearance is generated from the reference image, while all other objects use the same random appearance archetype.}
  \label{fig:replace_object}
\end{figure*}

\subsection{Generating the archetypes}\label{sec:generatingTheArchetypes}
The GUI enables the user to select from preexisting object appearances, as well as copying the appearance vector $a_i$ from another image. The existing object appearances are given as 100 archetypes per object class. These are obtained, by applying the learned network $A$ to all objects in a given class in the training set and employing k-means clustering, in order to obtain 100 class means.

In the GUI, the archetypes are presented linearly along a slider. The order along the slider is obtained by applying a 1-D t-SNE~\cite{tsne} embedding to the 100 archetypes.

\subsection{Inferring the scene graph from the layout panel}

The GUI lets the users place objects on a schematic layout, see Fig.~\ref{fig:teaser}. Each object is depicted as a string in one of ten different font sizes, in order to capture the size element of $l_i$. The location in the layout determines the $5\times 5$ grid placement, which is encoded in the other part of $l_i$. Note, however, that the locations and sizes are provided as indications of the structure of the graph layout and not as absolute locations (or scene layout). The generating network maintains freedom in the object placements to match the semantic properties of the objects in the scene.

The coarse placement by the user is more intuitive and less laborious than specifying a scene graph. To avoid adding unwanted work for the users, the edge labels are inferred, based on the relative position and size of the objects. An object which is directly to the left of another object, for example, is labeled ``left of''. The order in which objects are inserted to the layout determines the reference directing. If object $i$ is inserted before object $j$, then ``$i$ is to the left of $j$'' and not ''$j$ is to the right of $i$''. The inside and surrounding relations are determined similarly, by considering objects of different sizes, whose centers are nearby.

\subsection{Training details}
All networks are trained using \textit{ADAM} \cite{adam} solver with $beta1= 0.5$ for 1 million iterations. The learning rate was set to $1e^{-4}$ for all components except $\mathcal{L}_\text{D-mask}$, where we set it to a smaller learning rate of $1e^{-5}$. The different learning rates help us to stabilize the mask network.
We use batch sizes of 32, 16, 4 in our $64\times64$, $128\times128$, $256\times256$ resolutions respectively. Notice that since each image contains up to 8 objects, each batch contains up to $8\times32=256$ different objects.

\section{Experiments}

We compare our results with the state of the art methods of~\cite{Johnson2018ImageGF} and~\cite{Zhao2018ImageGF}, using various metrics from the literature. In addition, we perform an ablation analysis to study the relative contribution of various aspects of our method. Our experiments are conducted on the COCO-Stuff dataset~\cite{Caesar2018COCOStuffTA}, which, using the same split as the previous works, contains approximately 25,000 train images, 1000 validation, and 2000 test images.

We employ two modes of experiments: either using the ground truth (GT) layout or the inferred layout. The first mode is the only one suitable for the method of~\cite{zhao2018multiple}. Whenever possible, we report the statistics reported in the previous work. Some of the statistics reported for~\cite{Johnson2018ImageGF} are computed by us, based on the published model.
We report results for three resolutions $64^2$, $128^2$, and $256^2$. The literature reports numerical results only for the first resolution. While~\cite{Johnson2018ImageGF} presents visual results for 128x128, our attempts to train their method using the published code on this resolution, resulted in sub-par performance, despite some effort. We, therefore, prefer not to provide these non-competitive baseline numbers. The code of~\cite{zhao2018multiple} is not yet available.

We employ multiple acceptable literature evaluation metrics for evaluating the generated images. The {\em inception score}~\cite{Salimans2016ImprovedTF} measures both the quality of the generated images and their diversity. As has been done in previous works, a pre-trained inception network~\cite{inception} is employed in order to obtain the network activations used to compute the score. Larger inception scores are better. The {\em FID}~\cite{NIPS2017_7240} measures the distance between the distribution of the generated images and that of the real test images, both modelled as a multivariate Gaussian. Lower {\em FID} scores are better.

Less common, but relevant to our task, is the {\em classification accuracy} score, used by~\cite{zhao2018multiple}. A ResNet-101 model~\cite{he2016deep} is trained to classify the 171 objects available in the training datasets, after cropping and resizing them to a fixed size of 224x224 pixels. On the test image, we report the accuracy of this classifier applied to the object images that are generated, using the bounding box of the image's layout. A higher accuracy means that the method creates more realistic, or at least identifiable, objects.

We also report a {\em diversity score}~\cite{perceptualMetric}, which is based on the perceptual similarity~\cite{Johnson2015ImageRU} between two images. This score is used to measure the distance between pairs of images that are generated given the same input. Ideally, the user would be able to obtain multiple, diverse, alternative outputs to choose from. Specifically, the activations of AlexNet~\cite{krizhevsky2012imagenet} are used together with the LPIPS visual similarity metric~\cite{perceptualMetric}. A higher diversity score is better. 

In addition, we also report three scores for evaluating the quality of the bounding boxes. The {\em IoU} score is the ratio between the area of the ground truth bounding box that is also covered by the generated bounding box (the intersection), and the area covered by either box (the union). We also report recall scores at two different thresholds. {\em R@0.5} measures the ratio of object bounding boxes with an IoU of at least 0.5, and similarly for {\em R@0.3}.

\begin{table}[t]%
\begin{small}
\begin{center}
\begin{minipage}{\columnwidth}
\begin{tabular}{|l|l|l|l|l|}
  \hline
  Reso- & Method & Inception\footnote{The inception score of~\cite{Johnson2018ImageGF} for the complete pipeline is taken from their paper. The other scores are not reported there. The inception score for~\cite{Zhao2018ImageGF} is the one reported by the authors.} & FID & Accu- \\
  lution\footnote{\cite{Johnson2018ImageGF} and~\cite{Zhao2018ImageGF} report numerical results only for a resolution of 64x64.}& & & & racy \\ \hline\hline 
  \multirow{5}{*}{64x64} & Real Images & $16.3\pm0.4$ & 0 & 54.5 \\ \cline{2-5}
  &\cite{Johnson2018ImageGF} GT Layout & $7.3\pm0.1$ & 86.5 & 33.9\\
  &\cite{Zhao2018ImageGF} GT Layout & $9.1\pm0.1$ & \footnote{Not reported and cannot be computed due to lack of code/results.} & \footnote{The accuracy reported by~\cite{Zhao2018ImageGF} is incompatible (different classifiers).}\\
  & Ours GT Layout & $10.3 \pm 0.1$ & 48.7 & 46.1\\ \cline{2-5}
  &\cite{Johnson2018ImageGF} & $6.7\pm0.1$ & 103.4 & 28.8 \\
  &Ours & $7.9\pm 0.2$ & 65.3 & 43.3\\
    \hline
  \multirow{3}{*}{128x128} & Real Images & 24.2$\pm$ 0.9 &0 & 59.3\\\cline{2-5}
  & Ours GT Layout & $12.5\pm 0.3$ & 59.5 & 44.6\\\cline{2-5}
  & Ours & $10.4\pm0.4$ & 75.4 & 42.8\\
  \hline
\multirow{3}{*}{256x256} & Real Images & $30.7 \pm 1.2$ &0 & 62.4\\\cline{2-5}
  &Ours GT Layout & $16.4 \pm 0.7$ & 65.2 & 45.3 \\\cline{2-5}
  &Ours & $14.5 \pm 0.7$ & 81.0 & 42.2 \\
  \hline
 \end{tabular}

\end{minipage}
\end{center}
\end{small}
\caption{A quantitative comparison using various image generation scores. In order to support a fair comparison, our model does not use location attributes and employs random appearance attributes.}
\vspace{-.4cm}
\label{tab:quantitative_results}
\end{table}%

\begin{table}
\begin{small}
    \begin{tabular}{|l|l|c|}
    \hline
    Res & Method & Diversity\\
    \hline\hline
    64x64 &     Johnson \etal \cite{Johnson2018ImageGF} &  $0.15\pm 0.08$\\
          &     Zhao \etal \cite{Zhao2018ImageGF} GT layout &  $0.15\pm 0.06$\\
        & Ours  fixed appearance attributes & $0.23 \pm 0.01$\\
           &~~~~~~ and zeroed location attributes & \\
        & Ours  zeroed location attributes & $0.35\pm 0.01$\\
        & Ours  fixed appearance attributes & $0.37\pm 0.01$\\  
        & Our  full method  & $0.43\pm 0.07$\\           
         \hline
         
    256x256    & Ours  fixed appearance attributes & $0.48 \pm 0.09$\\
           &~~~~~~ and zeroed location attributes & \\
         &Ours  zeroed location attributes & $0.61\pm 0.07$\\
         &Ours  fixed appearance attributes & $0.62\pm 0.05$\\  
         & Our  full method  & $0.67\pm0.05$\\  
         \hline
    \end{tabular}
    \end{small}
    \caption{The diversity score of~\cite{perceptualMetric}. The results of~\cite{Johnson2018ImageGF} are computed by us and are considerably higher than those reported for the same method by~\cite{Zhao2018ImageGF}. The results of~\cite{Zhao2018ImageGF} are from their paper.} 
    \label{tab:diversity}
\smallskip
\smallskip
\begin{small}
    \begin{minipage}{\columnwidth}
    \centering
    \begin{tabular}{|l|l|l|l|}
    \hline
    & IoU & R@0.5 & R@0.3\\
    \hline\hline
         Johnson \etal \cite{Johnson2018ImageGF}\footnote{Taken from the paper itself} & 0.37 & 0.32 & 0.52\\
         Ours (w/o location attributes) & 0.41 & 0.37 & 0.62\\
         Ours (w/ location attributes) & 0.61 & 0.66 & 0.86\\
         \hline
    \end{tabular}
    \end{minipage}
    \end{small}
    \caption{Comparison of predicted bounding boxes}
    \label{tab:boundingbox}
\smallskip
\begin{small}
    \begin{center}
\begin{tabular}{|l|c|c|}
    \hline   
    User Study & \cite{Johnson2018ImageGF}& Ours \\
    \hline\hline
         More realistic output & 16.7\% & 83.3\%\\
         Better adherence to scene graph & 19.3\% & 80.7\%\\
         Ratio of observed objects   & 27.31\% & 45.38\%\\
         \quad  among all COCO objects & & \\
         Ratio of observed objects  & 46.49\% & 65.23\%\\
         \quad  among all COCO stuff & & \\
         \hline
    \end{tabular}
    \end{center}
    \end{small}
    \caption{User study results}
    \label{tab:userstudy}
\end{table}
\begin{table}
\begin{small}
\begin{center}
\begin{tabular}{|l|c|c|}
  \hline
  Model & Inception & FID\\ \hline\hline
  Full method & $10.4\pm0.4$ & 75.4\\
  No $\mathcal{L}_\text{perceptual}$ & $6.2\pm0.1$ & 125.1\\
  No $\mathcal{L}_\text{D-mask}$  & $5.2 \pm 0.1$	& 183.6\\
  No $\mathcal{L}_\text{D-image}$  & $7.4\pm0.2$ &	122.5\\
  No $\mathcal{L}_\text{D-object}$ & $8.7\pm0.1$ & 94.5\\
  Using $D_\text{image}$ of \cite{Johnson2018ImageGF}& $8.1\pm0.3$ & 114.2\\
  \hline
\end{tabular}
\end{center}
\end{small}
\caption{Ablation Study}
\vspace{-.4cm}
\label{tab:ablation}
\end{table}

Tab.~\ref{tab:quantitative_results} compares our method with the baselines and the real test images using the inception, FID, and classification accuracy scores. We make sure not to use information that the baseline method of~\cite{Johnson2018ImageGF} is not using and use zero location attributes and appearance attributes that are randomly sampled (see  \ref{sec:generatingTheArchetypes}).~\cite{zhao2018multiple} employs bounding boxes and not masks. However, we follow the same comparison (to masked based methods) given in their paper. 

As can be seen, our method obtains a significant lead in all these scores over the baseline methods, whenever such a comparison can be made. This is true both when the ground truth layout is used and when the layout is generated. As expected, the ground truth layout obtains better scores.

Sample results of our 256x256 model are shown in Fig.~\ref{fig:comparison}, using test images from the COCO-stuff datasets.
Each row presents the scene layout, the ground truth image from which the layout was extracted, our method's results, where the object attributes present a random archetype and the location attributes are zeroed ($l_i=0$), our results when using the ground truth layout of the image (including masks and bounding boxes), our results where the appearance attributes of each object are copied from the ground truth image and the location vectors are zero, and our results where the location attributes coarsely describe the objects' locations and the appearance attributes are randomly selected from the archetypes. In addition, we present the result of the baseline method of~\cite{Johnson2018ImageGF} at the 64x64 resolution for which a model was published.

As can be seen, our model produces realistic results across all settings, which are more pleasing than the baseline method. Using ground truth location and appearance attributes, the resulting image better matches the test image.

Tab.~\ref{tab:diversity} reports the diversity of our method in comparison to the two baseline methods. The source of stochasticity we employ (the random vector $z_i$ used in Eq.~\ref{eq:2}) produces a higher diversity than the two baseline methods (which also include a random element), even when not changing the location vector $l_1$ or appearance attributes $a_i$. Varying either one of these factors adds a sizable amount of diversity. In the experiments of the table, the location attributes, when varied, are sampled using per-class Gaussian distribution that fit to the location vectors of the training set images.

Fig.~\ref{fig:adiversity} presents samples obtained when sampling the appearance attributes. In each case, for all $i$, $l_i=0$ and the object's appearance embedding $a_i$ is sampled uniformly between the archetypes. This results in a considerable visual diversity. Fig.~\ref{fig:ldiversity} presents results in which the appearance is fixed to the mean appearance vector for all objects of that class and the location attribute vectors $l_i$ are sampled from the Gaussian distributions mentioned above. In almost all cases, the generated images are visually pleasing. In some cases, the location attributes sampled are not compatible with a realistic image. Note, however, that in our method, the default value for $l_i$ is zero and not a random vector.

Tab.~\ref{tab:boundingbox} presents a comparison with the method of~\cite{Johnson2018ImageGF}, regarding the placement accuracy of the bounding boxes. Even when not using the location attribute vectors $l_i$, our bounding box placement better matches the test images. As expected, adding the location vectors improves the results.

The ability of our method to copy the appearance of an existing image object is demonstrated in Fig.~\ref{fig:replace_object}. In this example, we generate the same test scene graph, while varying a single object in accordance with five different options extracted from images unseen during training. Despite the variability of the appearance that is presented in the five sources, the generated images mostly maintain their visual quality. These results are presented at a resolution of 256x256, which is the default resolution for our GUI. At this resolution, the system processes a graph in 16.3ms.

\noindent{\bf User study} Following~\cite{Johnson2018ImageGF}, we perform a user study to compare with the baseline method the realism of the generated image, the adherence to the scene graph, as well as to verify that the objects in the scene graph appear in the output image. The user study involved $n=20$ computer graphics and computer vision students. Each student was shown the output images for $30$ random test scene-graphs from the COCO-stuff dataset and was asked to select the preferable method, according to two criteria: ``which image is more realistic'' and ``which image better reflects the scene graph''. In addition, the list of objects in the scene graph was presented, and the users were asked to count the number of objects that appear in each of the images. The two images, one for the method of~\cite{Johnson2018ImageGF} and one for our method, were presented in a random order. To allow for a fair comparison, the appearance archetypes were selected at random, the location vectors were set to zero for all objects, and we have used images from our $64\times64$ resolution model. The results, listed in Tab.~\ref{tab:userstudy}, show that our method significantly outperforms the baseline method in all aspects tested.

\noindent{\bf Ablation analysis} The relative importance of the various losses is approximated, by removing it from the method and training the 128x128 model. For this study, we use both the inception and the FID scores. The results are reported in Tab.~\ref{tab:ablation}. As can be seen, removing each of the losses results in a noticeable degradation. Removing the perceptual loss is extremely detrimental. 
Out of the three discriminators, removing the mask discriminator is the most damaging, since, due to the random component $z_i$, we do not have a direct loss on the mask. Finally, replacing our image discriminator with the one in ~\cite{Johnson2018ImageGF}, results in some loss of accuracy.

\section{Conclusion}

We present an image generation tool in which the input consists of a scene graph with the potential addition of location information. Each object is associated both with a location embedding and with an appearance embedding. The latter can be extracted from another image, allowing for a duplication of existing objects to a new image, where their layout is drastically changed. In addition to the dual encoding, our method presents both a new architecture and new loss terms, which leads to an improved performance over the existing baselines.

\section*{Acknowledgement}
This project has received funding from the European Research Council (ERC) under the European Unions Horizon 2020 research and innovation programme (grant ERC CoG
725974).

{\small
\bibliographystyle{ieee_fullname}
\bibliography{egbib}
}

\appendix

\section{Network architecture}

The graph convolutional network used is the same as the one used in~\cite{Johnson2018ImageGF}, which was modified in order to support the added node information. We concatenate the embedding of the objects to the location attributes creating vectors in $R^{128+35}$. These vectors, together with the relation embedding is feed-forward into a fully-connected layer which results in a vector embedding in $\mathbb{R}^{128}$ for each of the objects and each of the relations. The network then follows the architecture of~\cite{Johnson2018ImageGF}, using a shared symmetric function to calculate the object vector from all the relations it participate in. Our graph convolution has overall five layers. 

To describe the rest of the networks, we follow a semi-conventional shorthand notation for defining architectures. Let $C_k$ denote a Convolution layer of $k$ filters with a kernel size of $7x7$ and a stride of $1$, followed by instance normalization~\cite{ulyanov2016instance} and a ReLU activation function. Similarly, we use $D_k$ to a layer which uses a stride of 2, reflection padding, and $k$ filters. In addition, we use $B_k$ to denote a $3x3$ upsample-convolution-BatchNormalization-ReLU layer with $k$ filters and a stride and padding of $1$. We use $V_k$ to define Residual blocks with two $3x3$ convolutional layers, both with $k$ filters. Moreover, $U_k$ denotes a layer with $k$ filters of size $3x3$ and a fractional stride of $0.5$, followed by instance normalization.  $CB_k$ denotes a stride-2 k-filter, $4x4$ convolution followed by batch normalization. $GA$ denotes a global average pooling layer. Fully connected layers with $k$ hidden units followed by a ReLU activation are denoted by $L_k$. The ReLU is not applied to the $L_k$ layer, if it is the top layer. 

The discriminators call for an even more elaborate terminology. 
Let $C_{i-k-o}$ denote a 3x3 Convolution layer with $i$ input channels and $k$ output filters, a stride of $o$ and a padding of 1. In addition, $LR$ denotes Leaky-ReLU with negative slope of $0.2$, $IN$ denotes Instance Normalization, and $AP_k$ denotes a $3x3$ Average Pool stride $2$ and padding of $1$. Also, let $C_{k-s}$ denote a Convolution layer with $k$ filters, stride of $s$, kernel size of $4x4$, and a padding of size $2$. $D_{k-s}$ denotes a $4x4$ Convolution-InstanceNorm-LeakyReLU layer with $k$ filters, a stride of $s$ padding of $2$ and a LeakyReLU with a negative slope of $0.2$.

The different components of the networks can be described as:
\begin{itemize}
    \item [$M$] $B_{192}$, $B_{192}$, $B_{192}$, $B_{192}$, $B_{192}$, $B_{192}$, $C_1$, Sigmoid activation
    \item [$B$] $L_{128}$, $L_{512}$, $L_4$
    \item [$A$] $CB_{64}$, $CB_{128}$, $CB_{256}$, $GA$, $L_{192}$, $L_{64}$, $L_{32}$
    \item [$R$] $C_{c+32}$, $D_{128}$, $D_{256}$, $D_{512}$, $D_{1024}$, $V_{1024}$, $V_{1024}$, 
$V_{1024}$, $V_{1024}$, $V_{1024}$, $V_{1024}$, $V_{1024}$, $V_{1024}$, $V_{1024}$, 
$U_{512}$, $U_{256}$, $U_{128}$, $U_{64}$, $C_3$
\item [$D_\text{mask}$] $C_{1-64-2}$, $LR$, $C_{(c+64)-128-1}^*$, $IN$, $LR$, $C_{128-1-1}$, $AP$
\item [$D_\text{image}$]$C_{(c+32)-2}$, $LR$, $D_{64-2}$, $D_{128-2}$, $D_{256-1}$, $C_{512-1}$
\item [$D_\text{object}$] $CB_{64}$,$CB_{128}$,$C4_{256}$, $GA$, $L_{1024}$, $L_{1}$
\end{itemize}
($^*$Concatenating the $c_i$ data in $D_\text{image}$ is done at the third layer)

\end{document}